\documentclass[nohyperref]{article}

\usepackage{microtype}
\usepackage{graphicx}
\usepackage{subfigure}
\usepackage{booktabs} 
\usepackage{comment}
\usepackage{hyperref}

\newcommand{\ys}[1]{{\color{green}{YS: #1}}}

\usepackage[accepted]{icml2022}

\usepackage{amsmath}
\usepackage{amssymb}
\usepackage{mathtools}
\usepackage{amsthm}
\usepackage{thmtools}
\usepackage[tiny,compact]{titlesec}
\titlespacing*{\paragraph}{0pt}{1.25ex plus 1ex minus .2ex}{1em}

\usepackage{graphicx}
\usepackage{subfiles}
\usepackage[capitalize,noabbrev]{cleveref}

\theoremstyle{plain}
\newtheorem{theorem}{Theorem}[section]
\newtheorem{proposition}[theorem]{Proposition}
\newtheorem{lemma}[theorem]{Lemma}
\newtheorem{corollary}[theorem]{Corollary}
\theoremstyle{definition}

\theoremstyle{remark}

\usepackage[textsize=tiny]{todonotes}

\icmltitlerunning{Unified Fourier Perspective on Equivariant Networks}

\begin{document}

\twocolumn[
\icmltitle{Unified Fourier-based Kernel and Nonlinearity Design for Equivariant Networks on Homogeneous Spaces}

\icmlsetsymbol{equal}{*}

\begin{icmlauthorlist}
\icmlauthor{Yinshuang Xu}{equal,yyy}
\icmlauthor{Jiahui Lei}{equal,yyy}
\icmlauthor{Edgar Dobriban}{yyy,comp}
\icmlauthor{Kostas Daniilidis}{yyy}
\end{icmlauthorlist}

\icmlaffiliation{yyy}{Department of Computer and Information Science, University of Pennsylvania, Philadelphia, PA 19104, United States}
\icmlaffiliation{comp}{Department of Statistics and Data Science, University of Pennsylvania, Philadelphia, PA 19104, United States}

\icmlcorrespondingauthor{Yinshuang Xu}{xuyin@seas.upenn.edu}
\icmlcorrespondingauthor{Jiahui Lei}{leijh@seas.upenn.edu}
\icmlcorrespondingauthor{Edgar Dobriban}{dobriban@wharton.upenn.edu}
\icmlcorrespondingauthor{Kostas
Daniilidis}{kostas@cis.upenn.edu}

\icmlkeywords{Machine Learning, ICML}

\vskip 0.3in
]

\printAffiliationsAndNotice{\icmlEqualContribution}

\begin{abstract}
We introduce a unified  framework for group equivariant networks on homogeneous spaces derived from a Fourier perspective. 
We consider tensor-valued feature fields, before and after a convolutional layer. We present
 a unified derivation of  kernels via the Fourier domain by leveraging the sparsity of Fourier coefficients of the lifted feature fields. The  sparsity emerges when the stabilizer subgroup of the homogeneous space  is a compact Lie group. 
We further introduce a nonlinear activation, via an elementwise nonlinearity on the regular representation after lifting and projecting back to the field through an equivariant convolution. 
We show that other methods treating features as the Fourier coefficients in the stabilizer subgroup are special cases of our activation. Experiments on $SO(3)$ and $SE(3)$ show  state-of-the-art performance in spherical vector field regression, point cloud classification, and molecular completion.

\end{abstract}

\section{Introduction}
Following the success of convolutional neural networks (CNNs) ~\cite{fukushima1980neocognitron,lecun1989backpropagation}, 
researchers made great strides in designing equivariant networks for groups beyond the standard translation operation.
Equivariance preserves symmetries and drastically reduces sample complexity, making data augmentation unnecessary and consequently reducing training and testing time.

The standard approach for equivariance by design is group-convolutional networks (G-CNNs), specified by convolution kernels~\cite{cohen2016group}. ~\citet{kondor2018generalization} 
prove that such group convolutions are both sufficient and necessary for linear layers to be equivariant to the action of compact groups. 
Existing group convolutional nets construct equivariant kernels in a case by case fashion; for example,  ~\citet{cohen2018general} present a general constraint for the kernel of G-CNNs with features on a homogeneous space that is solved algebraically, while ~\citet{finzi2021practical} solve for kernels numerically based on a similar finite-dimensional constraint. 

We propose a unified recipe for kernel design.
We lift the feature fields from the initial homogeneous space to the corresponding Mackey functions on the acting group, as introduced in~\cite{cohen2018general}.
Since a Mackey function satisfies a well-known constraint, it has redundant information. Specifically, when the irreducible representation  of the stabilizer subgroup is trivial, ~\citet{kondor2018generalization} proves that the Fourier Transform of the Mackey function has a certain block sparsity pattern. 
This naturally leads to the question whether the Fourier transform of Mackey functions for nontrivial irreducible representations over more general groups may have a similar property. 
We prove that when the stabilizer subgroup is a compact Lie group, the Fourier coefficients are sparse and  nonzero for specific field types, as stated in proposition \ref{Fourier sparse}. 
This spectral sparsity, appearing in both the input and the output of the convolution,
implies that the kernel itself can be taken as sparse.
This enables us to characterize the space of kernels without using the classical equivariance constraints and lays 
 the foundation of our framework for linear layer design.

Regarding the nonlinearity,  most existing  group convolution methods on  homogeneous spaces ~\cite{cohen2018general} apply norms or gated nonlinearities. The exceptions are works applying the Clebsch-Gordan decomposition of tensor products ~\cite{kondor2018clebsch} or interpreting the features as Fourier coefficients ~\cite{de2020gauge, poulenard2021functional}, since the feature vectors are in the vector  space under the irreducible representation of the stabilizer subgroup.
We propose a general formulation for nonlinear layers that consists of lifting the fields to the Mackey functions on the group, applying an elementwise nonlinearity, and finally projecting back to a field on the homogeneous space through a convolution. We prove the equivariance of this activation, and  that the recently proposed equivariant nonlinearity in ~\citet{poulenard2021functional} is a special case. 

In summary, our contributions are: 

1. We provide a unifying Fourier view of group convolution on homogeneous spaces dealing with different field types, 
and we prove
 that the input, output, and the kernel of the convolution are block-sparse in the Fourier domain when the stabilizer subgroup is a compact Lie group. Given the irreducible representations of the group, this harmonic view leads to an efficient method for designing linear group convolution layers. 
 
2. We present a novel approach to equivariant nonlinearities by applying elementwise nonlinearities to the feature fields lifted to Mackey functions on the group. 
We propose and implement a general approach for projecting functions on the group to fields on the homogeneous space in the final step of the nonlinearity.

The main goal and novelty of the paper is to use the Fourier coefficients for a unified derivation of kernels and nonlinearities.  We reach state-of-the-art results in standard equivariance benchmarks in 3D shape classification, molecular completion, and spherical vector field regression.
\section{Related Work}
\paragraph{Equivariant Networks}

The most common method to design equivariant networks is via group convolution, on the group or on the homogeneous space where the features lie. 

Existing work constructs group convolutional networks on images ~\cite{cohen2016group, worrall2017harmonic}, point clouds ~\cite{thomas2018tensor, chen2021equivariant}, voxel grid ~\cite{weiler20183d,worrall2018cubenet}, graphs ~\cite{maron2018invariant, keriven2019universal}, spherical images ~\cite{cohen2018spherical,esteves2018learning,esteves2020spin, cobb2020efficient} and sets ~\cite{maron2020learning,esteves2019equivariant}.
The above works that perform the group convolution directly on the homogeneous space implicitly lift the function to the group before the convolution, and project back to the homogeneous space after the convolution. 

Another way to achieve equivariance is through averaging over the group orbits ~\cite{puny2021frame, atzmon2021frame}. 
~\citet{finzi2020generalizing} proposed a general method for any Lie group with a surjective exponential map. 
For regular groups, \citet{bekkers2019b,sosnovik2019scale} propose a direct construction for group convolution. 
Recently, equivariance was introduced for attention networks ~\cite{fuchs2020se,romero2020group,hutchinson2021lietransformer,satorras2021n}. ~\citet{cohen2018general} and ~\citet{kondor2018generalization} showed for homogeneous spaces and compact groups, respectively, that an equivariant map can always be written as a convolution. ~\citet{finzi2021practical} introduced a numerical algorithm for equivariant linear layers based on solving linear systems involving the generators of the Lie algebra. ~\citet{weiler2021coordinate} equivariant maps on homogeneous spaces are convolutions with steerable kernels from differential geometriy perspective. ~\citet{aronsson2022homogeneous} generally analyzes the group convolutional neural networks for unimodular Lie groups with compact stablizer.
We  provide a general spectral approach for the theoretical analysis, and efficient implementations of equivariant networks on homogeneous space.

\paragraph{Equivariance and Fourier Transform}
Several works use the relationship between group convolutional networks and the Fourier Transform. 
~\citet{kondor2018generalization} provide a Fourier view of the group convolution for a scalar field on the quotient space, or on the whole group. 
~\citet{cohen2018spherical} and \citet{esteves2018learning} apply group convolutions on the Fourier transform of $SO(3)$ and the spherical harmonics, respectively, while 
~\citet{kondor2018clebsch} directly convolve and apply a non-linear activation in the spectral domain through the Clebsch-Gordan decomposition, using the compactness of $SO(3)$. ~\citet{esteves2020spin} deals with both vector and scalar signals on the sphere through the spectral domain, and is a special case of our general framework. 
Recently, ~\citet{cesa2021program} obtain explict general steerable filters for convolution through harmonic analysis for the stabilizer group.

\paragraph{Equivariant Nonlinearity}
It is nontrivial to design an  expressive and equivariant nonlinear layer, since an equivariant activation has to commute with the group action. Several works ~\cite{cohen2016group,cohen2018spherical,worrall2018cubenet} lift the signals from the homogeneous space to the group, and apply an elementwise activation to the group function. However, these methods deal with scalar fields and obtain the invariant features through global pooling without projecting back to the homogeneous space. In~\cite{thomas2018tensor,weiler20183d,worrall2017harmonic,esteves2020spin}  a nonlinearity is applied over invariant features such as norms. Because the direction of the tensor field remains unchanged, such a nonlinearity is not expressive enough, as explained in ~\cite{poulenard2021functional}.
~\citet{kondor2018clebsch} and \citet{anderson2019cormorant} apply a polynomial activation that can result in training instability, as mentioned in ~\cite{anderson2019cormorant}.
~\citet{weiler2019general} introduce nonlinearities with respect to various representations of $E(2)$, while 
~\citet{deng2021vector} generalize the classical ReLU activation to vectors (i.e., order-one tensors) by truncating an equivariant projection of the vector---a combination of a gated activation and a linear layer. ~\citet{de2020gauge}, \citet{poulenard2021functional} and \citet{cesa2021program} treat the features as Fourier coefficients, applying the Inverse Fourier Transform to functions on the stabilizer subgroup, and projecting back to the features by the Fourier Transform. We prove that such a nonlinearity is a special case of our method. 

The  works most closely related to ours are ~\cite{cohen2018general}, ~\cite{kondor2018generalization} and ~\cite{cesa2021program}. The main difference is that we provide a Fourier perspective for the group convolution on the homogeneous space, enabling efficient kernel design. 
Moreover, we provide a new nonlinearity and implement a general learnable equivariant linear map from the regular representation to the induced representation. 
~\citet{kondor2018generalization} analyze G-CNNs through Fourier analysis, but only focus on scalar fields. In contrast, we consider all field types (scalars, vectors, tensors) and our conclusion is consistent with ~\cite{kondor2018generalization} for the special case of scalar fields. 
~\citet{cesa2021program} solve the constraint of steerable filters in convolution by harmonic analysis on the stabilizer group, while we rely on the harmonic analysis for the full group and convolution theorem, which makes our approach easier to implement for non semidirect product groups.

\section{Method}

We give in App.~\ref{appendix:twist} the basic definitions that we need for a group acting  on a homogeneous space, as well as the definitions of cosets, bundles, fibers, and twist functions. In App.~\ref{appendix:irrep} we describe irreducible representations, and leave here only the tools we need for modeling features as tensor fields (Sec.~\ref{induced_and_mackey}) and the Fourier Transform (Sec.~\ref{fourier_transform}). Our work proposes two components for equivariant neural networks: the linear group convolution layer (Sec.~\ref{kernel_design})  and the nonlinear activation layer (Sec.~\ref{nonlinear_sec}).

\subsection{Induced representation and Mackey Functions}
\label{induced_and_mackey}
For a feature (or field) $f: G/H \rightarrow V$ over the homogeneous space  $G/H$  taking values in the vector space $V$, 
its type (scalar, vector or higher order tensor) is determined by an irreducible representation of the stabilizer subgroup $H$.  
We use $\rho$ to denote the unitary irreducible representation of $H$. Then $g\in G$ acts on the field as
\begin{align}
(\mathcal{L}_gf)(x) = \rho(\text{h}(g^{-1},x)^{-1})f(g^{-1}x)
\label{induced}
\end{align}
where 
$\text{h}$ is the twist function introduced in App.~\ref{appendix:twist}.
For example, for $G=SE(2)$, where $H=SO(2)$, 
the stabilizer subgroup has irreducible representations $\rho^m:SO(2)\mapsto \mathbb{C}$ for $m \in \mathbb{Z}$, given by $\rho^m(\theta)= e^{im\theta}$, for $\theta \in SO(2)$.
For scalar fields, we have $\rho(\theta)=\rho^0(\theta)=1$; for
vector fields, we have $\rho(\theta)=\rho^1(\theta)=e^{i\theta}$; and for a physical quantity like momentum of inertia over $\mathbb{R}^2$, we have 
$\rho(\theta)=\rho^2(\theta)=e^{2i\theta}$. In general, equation \eqref{induced} describes the action of $G$ on fields and $\mathcal{L}$ is called the induced representation, denoted as $\mathcal{L}=\text{Ind}^G_H\rho$.

The field $f:G/H\longrightarrow V$ can be lifted to a function
$f\!\!\uparrow^G:G \longrightarrow V $ on the group $G$  through an isomorphism $\Lambda$, via  
\[
    f\!\!\uparrow^G(g)=(\Lambda f)(g) = \rho (\text{h}(g)^{-1})f(gH) 
    \]
    and projected back via $\Lambda^{-1}$
    \[
     f(x)= f(s(x)H)= (\Lambda^{-1} (f\!\!\uparrow^G))(s(x)H)=f\!\!\uparrow^G(s(x)),
\]
where $\text{h}$ is the twist function 
and $s$ is the section map
(App.~\ref{appendix:twist}).
For the lifting in \citet{kondor2018generalization}, $\rho$ is a trivial representation, since only the scalar field is considered.
The action $\mathcal{L}'=\rho^G_{\text{reg}}$ of $G$ on $f\!\!\uparrow^G$ is a regular representation 
$(\mathcal{L}'_g(f\!\!\uparrow^G))(k)= f\!\!\uparrow^G(g^{-1}k)$ (~\citet{folland2016course}, Ch.~3). 

The lifting operation commutes with the group action, so  
$\rho^G_{\text{reg}}\circ \Lambda =\Lambda \circ \text{Ind}^G_H\rho$, 
which justifies calling the lifting an isomorphism.
The function $f\!\!\uparrow^G: G \longrightarrow V$ lifted from the field $f: G/H \longrightarrow V$
is called a Mackey function, satisfying
$f\!\!\uparrow^G(gh)=\rho(h^{-1})f\!\!\uparrow^G(g)$. 

\subsection{Fourier Transform}
\label{fourier_transform}
It is  known that the Fourier Transform exists for most groups of interest for equivariance: finite groups, compact groups, separable unimodular locally compact groups of Type I (see \citet{gross1978evolution}, \citet{folland2016course}, Ch.~7, and App.~\ref{appendix:def} for definitions), and certain semidirect product groups with an Abelian normal subgroup, like $SE(n)$ \cite{gauthier1991motions}. 
In these cases, there is a uniform (Haar) measure on the group $G$ that is both a left and a right Haar measure; we will denote its differential by $dg$.
Let $U(\cdot, p)$ be the unitary irreducible representation for each element $p \in \hat{G}$, where $\hat{G}$ is the dual of the group $G$, the set of equivalence classes of unitary irreducible representations of $G$ (\citet{folland2016course}, Ch.~7, \citet{chirikjian2001engineering}, Ch.~8). 
In the cases of interest, there is also a Haar measure $\nu$ on $\hat G$.
For any function $f:G \rightarrow \mathbb{C}$ that is square integrable with respect to the measure, the Fourier Transform of $f$ is defined as 
\begin{align*}
    \hat{f}(p)=\mathcal{F}(f)(p)= \int_G f(g)\overline{U(g,p)}dg
\end{align*}
while
the inverse Fourier Transform reads
\begin{align*}
    f(g)=\int_{\hat{G}} \textnormal{tr}(\hat{f}(p)^T U(g,p))d \nu(p)
\end{align*}
and the convolution and Parseval theorems both hold for $f \in \mathcal{L}^2(G)$. For vector valued functions, we apply both transforms component-wise.

\subsection{Unified Kernel Derivation }
\label{kernel_design}
\begin{figure*}
\centering
\includegraphics[width=6in]{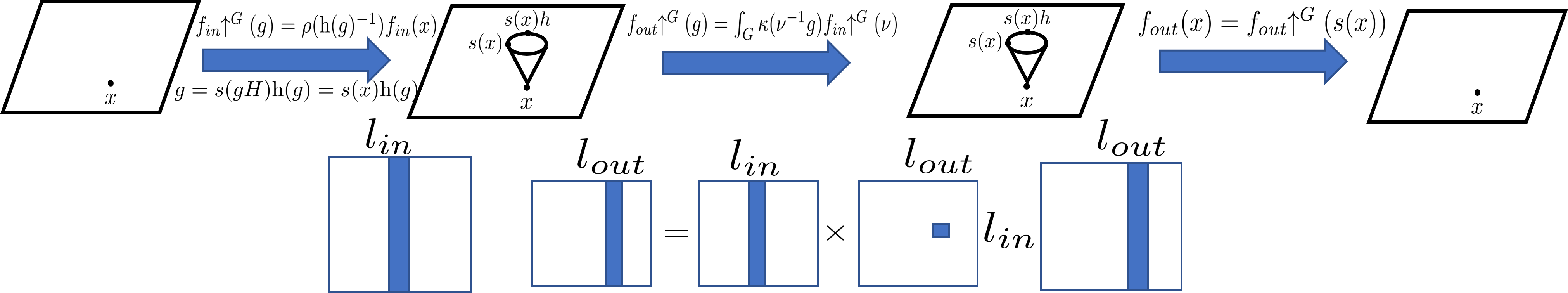}
\caption{\textbf{Convolution layer}: 
(a). Lifting the input field $f_{in}: G/H \rightarrow V$ to the Mackey function $f_{in}\!\!\uparrow^G: G \rightarrow V$. 
(b). Convolution with designed kernel to find a new Mackey function $f_{out}\!\!\uparrow^G: G \rightarrow V$ of order $l_{out}$. 
(c). Projection to the homogeneous space to find $f_{out}$. 
We highlight the sparsity in the spectral domain
of the input, kernel and output. 
}
\label{convolution_part}
\end{figure*}

As described in \citet{cohen2018general}, an equivariant linear layer, consisting of a convolution  with a kernel, can be realized in three steps: 
1) lifting the input field (feature map) $f_{in}: G/H \rightarrow V$ to the corresponding Mackey function  $f_{in}\!\!\uparrow^G: G \rightarrow V$ on the group; 
2) applying group convolution to the lifted function $f_{in}\!\!\uparrow^G$ with the kernel $\kappa$ to find the output Mackey function $f_{out}\!\!\uparrow^G$; 
3) projecting $f_{out}\!\!\uparrow^G$ back to the homogeneous space to obtain the output field (feature map) $f_{out}$. These three steps can be summarized as the following linear map yielding
\begin{align*}
\label{group}
    f_{out}(x) &=
    (\Lambda_2^{-1}(\kappa \ast (\Lambda_1 f_{in})))(x)\notag\\
    &= \int_G \kappa(g^{-1}s(x))(\Lambda_1 f_{in})(g)dg,
\end{align*}
where $\Lambda_1$ and $\Lambda_2$ are the lifting isomorphisms for the input and output fields, respectively, and $\kappa: G \rightarrow \text{Hom}(V_1,V_2)$.
When $G$ is a semidirect product, convolution simplifies to $$(\Lambda_2^{-1}(\kappa\ast (\Lambda_1 f)))(x) = \int_{G/H} \kappa'(s(x)^{-1}y)f(y)dy,$$ 
where $\kappa'(x)=\kappa(s(x))$. 
Previous work \cite{cohen2018general} derives the constraints for the kernel $\kappa$ when the input and output are Mackey functions by working in the spatial domain. In contrast, we analyze and design the linear layer from the spectral perspective. In  section \ref{fourier_subsec}, we will prove that the Fourier coefficients of a Mackey function obey certain sparsity  patterns, which also imply the sparsity of the kernel $\kappa$. 
Moreover, in certain cases our proposition leads to a more efficient implementation of convolutions in the spectral domain (compared to the spatial domain), as shown in Figure \ref{convolution_part}.

\subsubsection{Sparsity of Mackey functions in the Fourier domain}
\label{fourier_subsec}
In this section we prove that the Fourier transform of a Mackey function $f\!\!\uparrow^G$ lifted from the field $f$ has a certain sparsity pattern. To prove this proposition, we first need to introduce a lemma about unitary irreducible representations of $G$.

 \begin{lemma}
 \label{decomp}
 Let $G$ be a unimodular group, and its stabilizer subgroup $H$ be a compact Lie group. For any $p \in \hat{G}$, the dual group, the restricted representation for the unitary irreducible representation $U(\cdot,p)$, $U(\cdot,p)|_H$, can be decomposed as the direct sum $\oplus_{i \in \mathcal{Q}(p)}\rho^i(\text{h}(g))$, for an index set $\mathcal{Q}(p)$ parametrized by $p \in \hat{G}$, where $\rho^i$ are the irreducible representations of $H$. 
 \end{lemma}
See App.~\ref{appendix:lemma} for the proof.
Using Lemma~\ref{decomp} we can now describe the sparsity of Mackey functions in the Fourier domain.

\begin{proposition}
\label{Fourier sparse}
Assume $G$ is a unimodular group and its stabilizer subgroup is a compact Lie group. A Mackey function $f\!\!\uparrow^G: G\rightarrow V$  lifted via $f\!\!\uparrow^G(g)=\rho(\text{h}(g)^{-1})f(gH)$ from a field $f: G/H \rightarrow V$ has the following sparsity pattern in the Fourier domain:   $\left[\widehat{f\!\!\uparrow^G}\right](p)_{\star,j}$ is  nonzero only if the block at column $j$ in the decomposition of $U(\cdot,p)|_H$ is equivalent to the dual representation of $\rho$.
\end{proposition}

See App.~\ref{appendix:prop} for the proof.
Next, we show that this sparsity carries on to the convolution, ensuring that the spectrum of the kernel is also sparse. 

\begin{corollary}
\label{ker-sparse}
Let $f_1\!\!\uparrow^G$ and $f_2\!\!\uparrow^G$ be Mackey functions lifted from the fields $f_1$ and $f_2$, and $\rho_1$ and $\rho_2$ be the irreducible representations that determine the field type of $f_1$ and $f_2$.
For any group convolution $f_2\!\!\uparrow^G (g)=\int_G\kappa(\nu^{-1}g)f_1\!\!\uparrow^G(\nu)d\nu$,
without loss of generality, the kernel $\kappa$ has the following sparsity pattern on  its Fourier coefficients: $\left[\hat{\kappa}\right](p)_{i,j}$ is zero when the block at row $i$ in the decomposition of $U(\cdot,p)|_H$ is not equivalent to the dual representation of $\rho_{1}$, or the block at column $j$ in the decomposition of $U(\cdot,p)|_H$ is not equivalent to the dual representation of $\rho_{2}$.
\end{corollary} 

See App.~\ref{appendix:coro} for the proof.
In particular, when $\rho$ is the trivial representation, i.e., the field is  scalar-valued, and $G$ is compact, proposition \ref{Fourier sparse} recovers the conclusion of \cite{kondor2018generalization}.
In summary, we proved that the spectrum of the Mackey function and the corresponding kernel are sparse. This sparsity  enables us to directly and analytically design  kernels and implement the group convolution. 

Meanwhile, we state that our design gives a complete characterization of the space of kernels for equivariant convolutions. 
Given an appropriate unitary irreducible representation, the block sparsity stated in Prop. ~\ref{Fourier sparse} can be simplified to column sparsity. 
The Fourier coefficients of elements of vector functions are related. 
The converse also holds, 
hence the Fourier coefficients of a function $f$ have such sparsity and related values if and only if $f$ is a Mackey function. 
Further, with the convolution theorem, we find the sparsity in the spectrum of the kernel $\kappa$, and 
the relation of Fourier coefficients for every element in the matrix function. 
Finally, we prove that 
there is a bijection between the kernel space $\left\{\kappa:G \rightarrow \text{Hom}(V_1,V_2)\right\}$, where $\kappa(gh)=\rho_{out}(h^{-1})\kappa(g)$ and 
$\kappa(hg)=\kappa(g)\rho(h^{-1})$ for any $g \in G, h \in H$, and the kernel space with such a spectrum. See App.~\ref{appendix:discussion of completeness} for details. To help the reader, now explain the sparsity pattern for $SO(3)$ and $SE(2)$.

\begin{figure*}
\centering     
\subfigure[$SO(3)$]{\label{so3_mul}\includegraphics[width=3in]{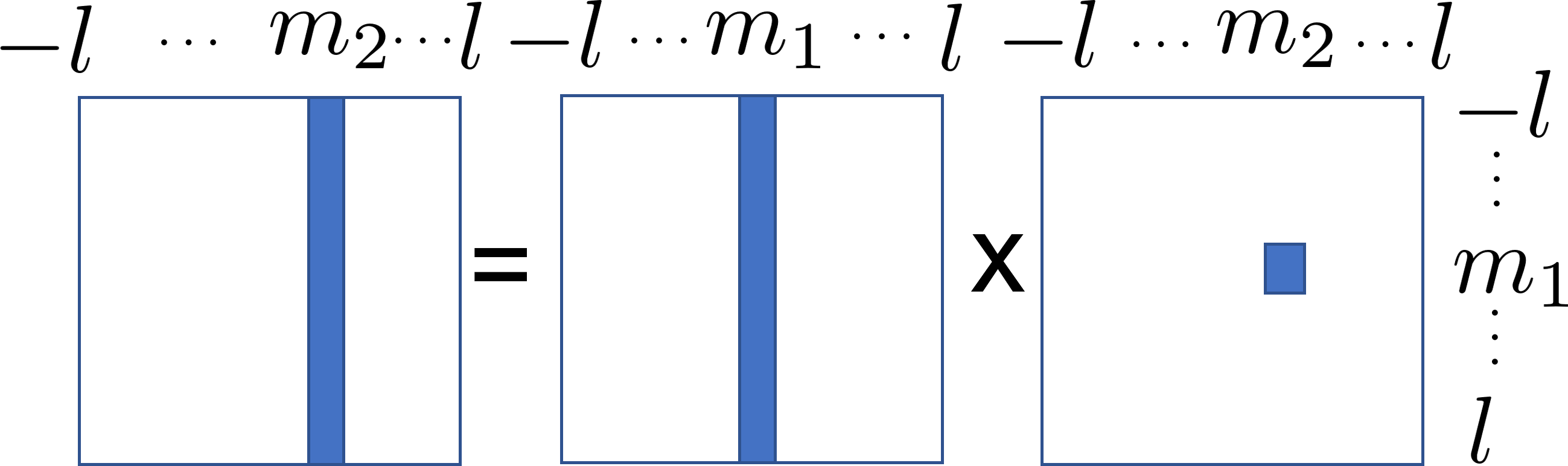}}
\subfigure[$SE(2)$]{\label{se2_mul}\includegraphics[width=3in]{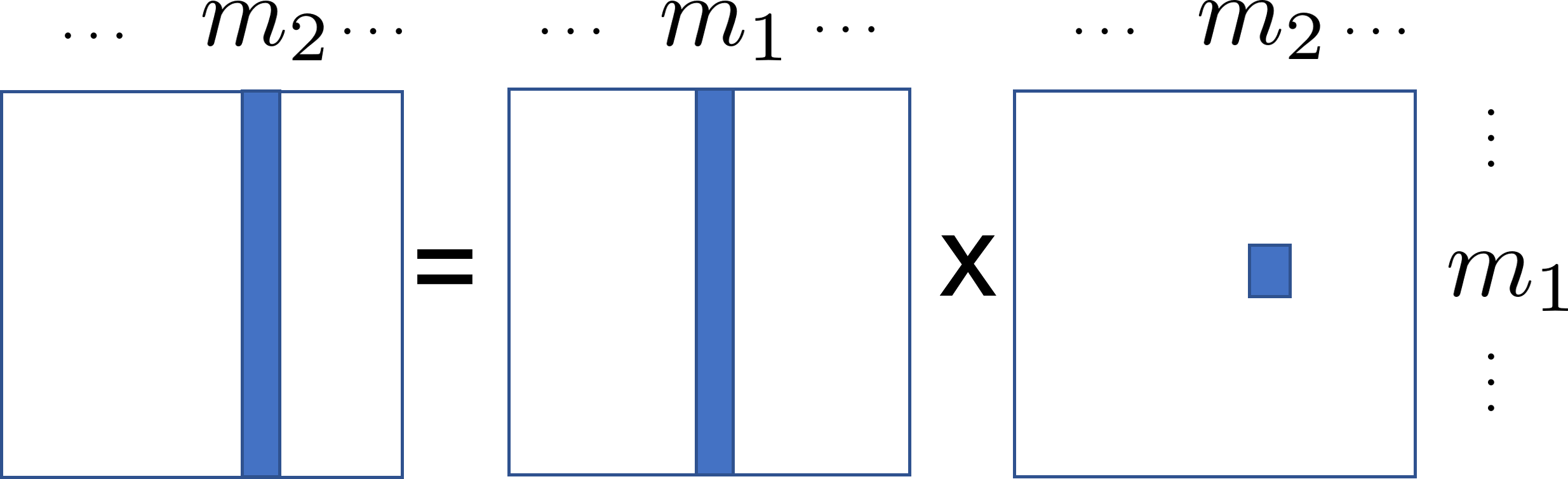}}
\caption{Convolution on $SO(3)$ and $SE(3)$ in the spectral domain.
We illustrate the sparsity of the Fourier coefficients of the input, output and kernel.}
\end{figure*}

\paragraph{Sparsity in the SO(3)-spectrum}
\label{example_so3}
Consider the group $G=SO(3)$, with the stabilizer subgroup $SO(2)$ and the homogeneous space $\mathbb{S}^2$.
A rotation matrix $R\in SO(3)$ can be parametrized by the Euler Z-Y-Z angles $(\alpha, \beta, \gamma)$. 
The unitary irreducible representations of  $SO(3)$ are indexed by integers $l$ and have the form
\begin{align*}
    D^l_{mn}(R)=D^l_{mn}(\alpha, \beta, \gamma)= e^{-im\alpha}d^l_{mn}(\beta)e^{-in\gamma},
\end{align*}
where $m,n$ are row and column indices, 
and $d^l_{mn}$ are the elements of Wigner's small $d$-matrices.
The lifting process and its inverse take the form
\begin{align*}
    &f\!\!\uparrow^G(R) = f\!\!\uparrow^G(\alpha, \beta, \gamma)= e^{-ik\gamma} f(\alpha, \beta), \\
    &  f(\alpha, \beta)= f\!\!\uparrow^G(\alpha, \beta, 0),
\end{align*}
where $k$ is the corresponding order of the irreducible representation of the field.
The Fourier Transform $\widehat{f\!\!\uparrow^G}^l_{mn}$ (see App.~\ref{appendix:$SO(3)$_detail} for details) becomes:
\begin{align*}
    \int_{(\alpha, \beta)\in \mathbb{S}^2} f(\alpha, \beta)e^{im\alpha}d^l_{mn}(\beta)d\alpha \sin(\beta) d\beta*2\pi\delta(k-n),
\end{align*}
where $\delta(x)$ is the Kronecker delta function, which is zero except at $x=0$, where it equals unity.
The convolution on $SO(3)$ takes the form 
$
(l_1\ast l_2) (g) = \int_{k\in G} l_1(\nu^{-1}g)l_2(\nu) d\nu
$
and the convolution theorem states
$\mathcal{F}(\kappa\ast f)^l_{mn}=\sum_{j}\hat{f}^l_{mj}\hat{\kappa}^l_{jn}$.
When the output field corresponds to an $m_2$-th order irrep of $SO(2)$ and the input field to an $m_1$-th order irrep, as shown in Figure \ref{so3_mul}, we find the following sparsity  structure of $\kappa$
\begin{align*}
  \hat{\kappa}^l_{mn}=\hat{\kappa}^l_{mn}\delta(m-m_1)\delta(n-m_2).
\end{align*}
By applying the inverse Fourier Transform, we find the kernel $\kappa(R)$ 
to be:
\begin{equation*}
  \sum^{\infty}_{l=0} c^l D^l_{m_1m_2}(R)
  = e^{-im_1\alpha} \,\sum^{\infty}_{l=0} c^l d^l_{m_1m_2}(\beta)\,\,e^{-im_2\gamma}.
\end{equation*}
Projecting the output Mackey function to the output field leads to (App.~\ref{appendix:$SO(3)$_detail})
\begin{align*}
    &f_{out}(\alpha,\beta)
    =C\int_{\mathbb{S}^2} e^{-im_2\text{h}(R^{-1}(
     \alpha',\beta',0)R(\alpha,\beta,0))}\\ 
     &\quad\quad\kappa'(R_{y}(-\beta')R_z(-\alpha')x_{\alpha,\beta})
     f_{in}(\alpha',\beta')d\alpha'sin(\beta')d\beta',
\end{align*} 
where $\kappa'$ is a function on the sphere and $\kappa'(\alpha, \beta)=\kappa(\alpha,\beta,0)$, $x_{\alpha,\beta}$ is point on the sphere and $C$ is a constant. When we apply the convolution on the homogeneous space $\mathbb{S}^2$, the twist $e^{-im_2\text{h}(R^{-1}(\alpha',\beta',0)R(\alpha,\beta,0))}$ appears, which is consistent with the correlation on the homogeneous space derived in \cite{cohen2018general} for non-semidirect product groups. This shows the difficulty of  implementing convolution on the homogeneous space directly in the spatial domain. Therefore, it is more efficient to implement the convolution in the spectral domain.
When the input and the output are scalar fields, the filter is isotropic, which is consistent with the results in \cite{esteves2017polar}.

\paragraph{Sparsity in the $SE(2)$-spectrum}
Consider the group $G=SE(2)=\mathbb{R}^2 \rtimes SO(2)$, where the homogeneous space is $\mathbb{R}^2$ and stabilizer subgroup is $SO(2)$. 
Any $g \in G$ can be parameterized as $(x,\theta)= (a,\phi,\theta)$, where we identify $x \in \mathbb{R}^2$ with its action $t_x$, etc; and where $a = |x|$ is the modulus of $x$, and for nonzero $a>0$, $x/a = e^{i\theta}$, for $\theta \in [0,2\pi)$. 
The unitary irreducible representations of $SE(2)$ are indexed by integers $v$ and $p \in \mathbb{R}^+$, and with $J_v$ denoting the $v$-th order Bessel function of the first kind, they have the form
\begin{align*}
    U_{mn}(g,p) = i^{n-m} e^{-i(n\theta + (m-n)\phi)}J_{n-m}(pa).
\end{align*}

The lifting process and its inverse is:
\begin{align*}
    &f\!\!\uparrow^G(a,\phi, \theta)=e^{-im\theta}f(a,\phi)\\
    & f(a,\phi) = f\!\!\uparrow^G(a,\phi,0), 
\end{align*}
where $m$ is the corresponding order of the irreducible representation of the field.
Then the Fourier Transform of a Mackey function $f\!\!\uparrow^G \in \mathcal{L}^2(G)$ is 
\iftrue
\begin{align*}
    &\widehat{f\!\!\uparrow^G}_{mn}=\int_{g \in G} f\!\!\uparrow^G(g) \overline{U_{mn}(g,p)}dg\\
     &=\int_{g \in G} e^{-ik\theta}f(a,\phi)i^{m-n}e^{i(n\theta+(m-n)\phi)}J_{n-m}(pa)dg\\
     &=\int_{(a,\phi) \in G/H}f(a,\phi) i^{m-n} J_{n-m}(pa) e^{i(m-n)\phi}\\&\quad\int_He^{-ik\theta}e^{in\theta}d\theta.
 \end{align*}
Further, $\widehat{f\!\!\uparrow^G}_{mn}$ equals
\begin{align*}
    \int_{(a,\phi) \in G/H} f(a,\phi) i^{m-n} J_{m-n}(pa) e^{i(m-n)\phi}*2\pi\delta(n-k).
\end{align*}
\fi
The convolution on $SE(2)$ has the form
$(l_1\ast l_2) (g) = \int_{k\in G} l_1(k^{-1}g)l_2(k) dk$.
The convolution theorem for $SE(2)$ states that
$\mathcal{F}(l_1\ast l_2)(p)=\hat{l}_2(p)\hat{l}_1(p)$,
which can be viewed as the multiplication of two infinite dimensional matrices. 

As shown in figure \ref{se2_mul}, when the input and output of the convolution are Mackey functions, the Fourier coefficients are nonzero on the $m_1$-th column and  $m_2$-th column 
Therefore, 
we obtain the kernel 
\begin{align*}
    \mathcal{F}(\kappa)(p)_{mn} =\hat{\kappa}(p)_{mn}\delta(m-m_1) \delta(n-m_2).
\end{align*}

By applying the inverse Fourier Transform,  
we find the kernel to be:
\begin{align*}
    \kappa(g)=\int_{0}^{\infty} c_p i^{m_2-m_1} e^{-i(m_2\theta + (m_1-m_2)\phi)}J_{m_2-m_1}(pa)pdp.
\end{align*}

Mackey functions can be projected to $\mathbb{R}^2$ to find
\begin{align*}
    \kappa'(x)&=\int_{0}^{\infty} c_{p} i^{m_2-m_1} e^{-i (m_1-m_2)\phi}J_{m_2-m_1}(pa)pdp\\
    &=e^{i(m_2-m_1)\phi}R(a),
\end{align*}
where $c_p$ is a constant, and $R$ is a radial function. 
This exactly recovers the form of the kernel in harmonic networks \cite{worrall2017harmonic}.

In App.~\ref{appendix: $SE(3)$_detail}, we provide another example, $SE(3)$; we have a similar analysis and conclusion. The block sparsity stated in Prop.~\ref{Fourier sparse} and Cor.~\ref{ker-sparse} also exist in  the $SE(3)$-spectrum for Mackey functions and the corresponding kernels. The derived kernels are consistent with the form of tensor field networks in \citet{thomas2018tensor} and 3D steerable CNNS in \citet{weiler20183d}.

\subsection{Equivariant nonlinearity}
\label{nonlinear_sec}

\begin{figure*}
\centering
\includegraphics[width=6in]{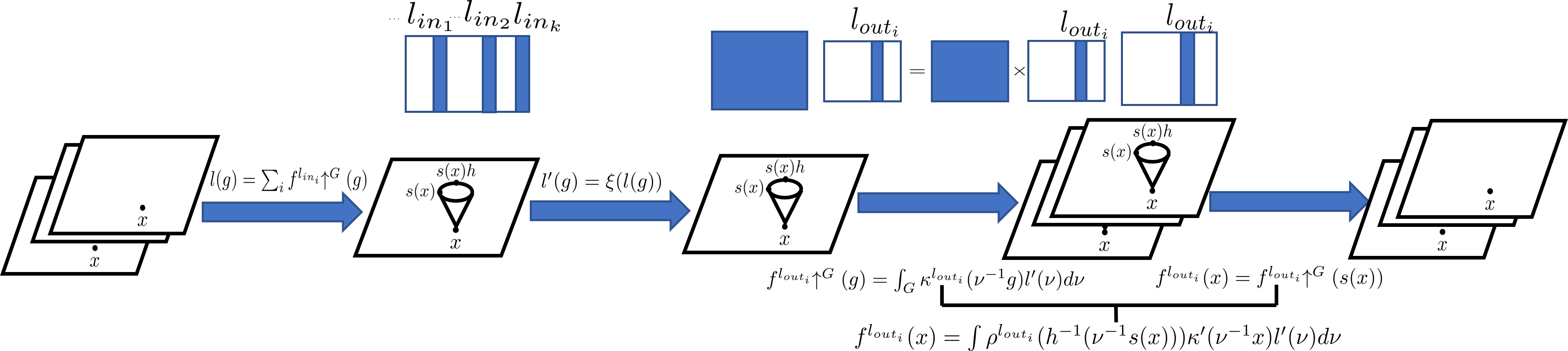}
\caption{\textbf{Activation layer}: (a). Lifting inputs 
$\left\{f^{l_{in_i}}:G/H \rightarrow V_{l_{in_i}}\right\}$
of different orders to the Mackey functions $\left\{f^{l_{in_i}}\!\!\uparrow ^G: G \rightarrow V_{l_{in_i}} \right\}$ and summing them up to find $l: G \rightarrow V$. 
(b). Element-wise activation on $l$. 
(c). Convolution with the designed kernel to find Mackey functions $\left\{f^{l_{out_i}}\!\!\uparrow ^G: G \rightarrow V_{l_{out_i}} \right\}$ of different orders. 
(d). Projection to the homogeneous space to find fields $\left\{f^{l_{out_i}}:G/H \rightarrow V_{l_{out_i}} \right\}$ of different orders.   
}
\label{nonlinear_part}
\end{figure*}

\label{acti}
The whole feature map consists of fields of different types, i.e., it can be written as $f(x)=\bigoplus_i f^{l_i}(x)$ where $l_i$ is the type of the field. 
Therefore, the group action on a feature map is the direct sum of the associated induced representations, $\bigoplus_i \text{Ind}^G_H \rho^{l_i}$. For an  equivariant nonlinearity $\sigma : V^{in} \rightarrow V^{out}$, we need
\begin{align}
    \bigoplus_j \text{Ind}^G_H \rho^{l_{out_j}}\circ \sigma =\sigma \circ \bigoplus_i \text{Ind}^G_H \rho^{l_{in_i}}.
\end{align}

The most common strategy to obtain equivariant nonlinear maps in modern deep learning is to apply a fixed nonlinear function $\xi$ elementwise to each coordinate of the feature $f$ in some fixed basis.
In general, this does not satisfy the above equivariance condition. 
However, this nonlinearity is equivariant for the regular representation, i.e., $\rho^G_{reg}(g)\circ\xi=\xi\circ \rho^G_{reg}(g)$, as mentioned in \citet{cohen2016group}: 
\begin{align*}
  &\left[ \rho^G_{reg}(g)\circ\xi (l)\right](k)=\left[\xi(l)\right](g^{-1}k)=\xi(l(g^{-1}k))\\
  &=\xi(\left[\rho^G_{reg}(g)(l)\right](k))=\left[\xi\circ \rho^G_{reg}(g)(l)\right](k),
 \end{align*}
 where $l:G \rightarrow V$.
Therefore, we can leverage the lifting isomorphism $\Lambda$, which satisfies that $\rho^G_{\text{reg}}\circ \Lambda =\Lambda \circ \text{Ind}^G_H\rho$, to lift the features to the corresponding Mackey functions. When there are multiple types of features $\left\{f^{l_i}: G/H \rightarrow V \right\}$, we simply apply the sum of these Mackey functions. We denote the composition of the lifting isomorphism and sum operation by $\overline{\Lambda}$, so that
\begin{align}
   &l(g)=\left[\overline{\Lambda}(\bigoplus f^{l_i})\right](g)= \sum_i \left[\Lambda_i(f^{l_i}) \right](g)\notag\\
   &=\sum_i f^{l_i}\!\!\uparrow^G\!\!(g)= \sum_{i}\rho^{l_i} (\text{h}(g)^{-1})f^{l_i}(gH),
\end{align}
where $\Lambda_i$ is the lifting isomorphism for the field $f^{l_i}$.
Clearly, $\rho^G_{\text{reg}}\circ \overline{\Lambda} =\overline{\Lambda} \circ \bigoplus_i \text{Ind}^G_H \rho^{l_i}$.
In general, $l$ is not a Mackey function, only a sum of different Mackey functions. Its spectral domain is shown in Figure \ref{nonlinear_part}.

After applying an elementwise nonlinearity to the group function, we need to project the group function to the homogeneous space. The signal may be any function on the group and its Fourier coefficients may be nonsparse, as shown in Figure \ref{nonlinear_part}. To maintain equivariance, we propose to convolve with a designed kernel as shown in Figure \ref{nonlinear_part} to find a Mackey function, and then project to the homogeneous space via
\begin{align*}
    f(x)&= \int \kappa(g^{-1}s(x))l(g)dg.
\end{align*}
In Figure \ref{nonlinear_part}, we know that the kernel $\kappa$ is also a Mackey function, satisfying $\kappa(gh)=\rho(h^{-1})\kappa(g)$.
Therefore $\kappa(g)$ can be expressed as $\kappa(g)=\rho(\text{h}(g)^{-1})\kappa(s(gH))=\rho(\text{h}(g)^{-1})\kappa'(gH)$, for $\kappa' = \kappa \circ s$. 
Then the convolution becomes: 
\begin{align*}
    f(x)&= \int \rho(\text{h}(g^{-1}s(x))^{-1})\kappa'(g^{-1}x)l(g)dg,
\end{align*}
where $\rho$ is the irreducible representation corresponding to the field $f$. Let us denote this projection by $P^{\kappa}$.
The convolution is equivariant, i.e., $\text{Ind}^G_H \rho \circ P^{\kappa} = P^{\kappa} \circ \rho^G_{reg}$.

Therefore, the nonlinearity $\sigma$ is the composition of lifting, elementwise nonlinearity and projection. We prove the equivariance of $\sigma$: 
 \begin{align*}
      &\sigma \circ \bigoplus_i \text{Ind}^G_H \rho^{l_{in_i}}=(\bigoplus_j P^{\kappa_{out_j}}) \circ \xi \circ \overline{\Lambda}\circ \bigoplus_i \text{Ind}^G_H \rho^{l_{in_i}}\notag\\
      &=(\bigoplus_j P^{\kappa_{out_j}}) \circ \xi \circ \rho^G_{reg}\circ\overline{\Lambda}\notag=(\bigoplus_j P^{\kappa_{out_j}}) \circ \rho^G_{reg} \circ \xi \circ \overline{\Lambda}\\
      &=\bigoplus_j \text{Ind}^G_H \rho^{l_{out_j}} \circ (\bigoplus_j P^{\kappa_{out_j}}) \circ  \xi \circ \overline{\Lambda}\\
      &=\bigoplus_j \text{Ind}^G_H \rho^{l_{out_j}} \circ \sigma,
 \end{align*}
 where $\kappa_{out_j}$ is the constrained kernel corresponding to $\text{Ind}^G_H\rho^{l_{out_j}}$. 

This is exactly the general form of non-linearity which treats the tensor fields on the homogeneous space as the Fourier coefficients. 
Taking $SE(3)$ as an example, suppose $G = SE(3)$ and $H= SO(3)$. We use the real form of the Wigner  $D$-matrix. 
Lifting multiple fields $\left\{f^l: \mathbb{R}^3 \rightarrow V\right\}$ to the function on the group has the form
\begin{equation*}
    l_n(x,R)= \sum^{l_{\max}}_{l=0} \sum_{m=-l}^lD^l_{nm}(R^{-1})f^l_m(x),
\end{equation*}
where $D^l$ is the Wigner  $D$-matrix, the irreducible representation for the field $f^l$.

When we take the 0-th element of $l: SE(3) \rightarrow V$, we find
\begin{align*}
\label{foward}
     &l_0(x,R)=  \sum^{l_{\max}}_{l=0} \sum_{m=-l}^lD^l_{0m}(R^{-1})f^l_m(x)\\
             &=  \sum^{l_{\max}}_{l=0} \sum_{m=-l}^lD^l_{m0}(R)f^l_m(x)
             =  \sum^{l_{\max}}_{l=0} \sum_{m=-l}^lD^l_{m0}(r)f^l_m(x)\\
             &=  \sum^{l_{\max}}_{l=0} \sum_{m=-l}^lY^l_m(r)f^l_m(x)
             =  \mathcal{F}^{+}(f(x))(r),
 \end{align*}
 
where $R=(\alpha, \beta, \gamma) \in G$  and $r=(\alpha, \beta) \in \mathbb{S}^2$. 
This is exactly the form of the signals on the sphere obtained in \cite{poulenard2021functional} 
through the inverse Spherical Harmonics Transform (iSHT).

On the other hand, when we take $\kappa(x)_{ij}=
\delta(i)\delta(j)\delta(x)$, the projection is equivalent to the Spherical Harmonics Transform (SHT) in \cite{poulenard2021functional}, as 
  \begin{align*}
    f^{l_i}_m(x)= &\sum_{n,j}\int D^l_{mn}(\text{h}(g^{-1}s(x))^{-1})\kappa_{nj}(g^{-1}x)\xi(l_j(g))dg\\
    =& C_1\int D^l_{m0}(R)\xi(l_0(x,R))dR\\
     =&C_2\int Y^l_m(r)\xi(\mathcal{F}^{+}(f(x))(r))dr,
 \end{align*}
  where $C_1$ and $C_2$ are constants. The second equality holds because $\kappa_{nj}$ can be nonzero only when $gH=x$, $n=0$ and $j=0$.

To keep the network simple, we do not use trainable weights for the kernel. Instead, we take $\kappa(x)_{ij}=\delta(x)\delta(i-j)$, where the first $\delta$ is the Dirac delta function and the second is the Kronecker delta function as used previously. Then the projection becomes $f_i(x)=\int_H \sum_j\rho_{ij}(h)l_j(s(x)h)dh$. \citet{weiler2019general} describe such a nonlinearity when the stabilizer subgroup $H$ is $O(2)$. In the spectral domain, we can extract the corresponding column from the Fourier matrix of $l(g)$ for simplicity.

\section{Implementation and Results}

\subsection{SO(3): Vector field prediction on the sphere}
\label{sec:exp_sph}
\begin{figure}[ht]
\centering
\scalebox{0.8}{
\centerline{\includegraphics[width=\columnwidth]{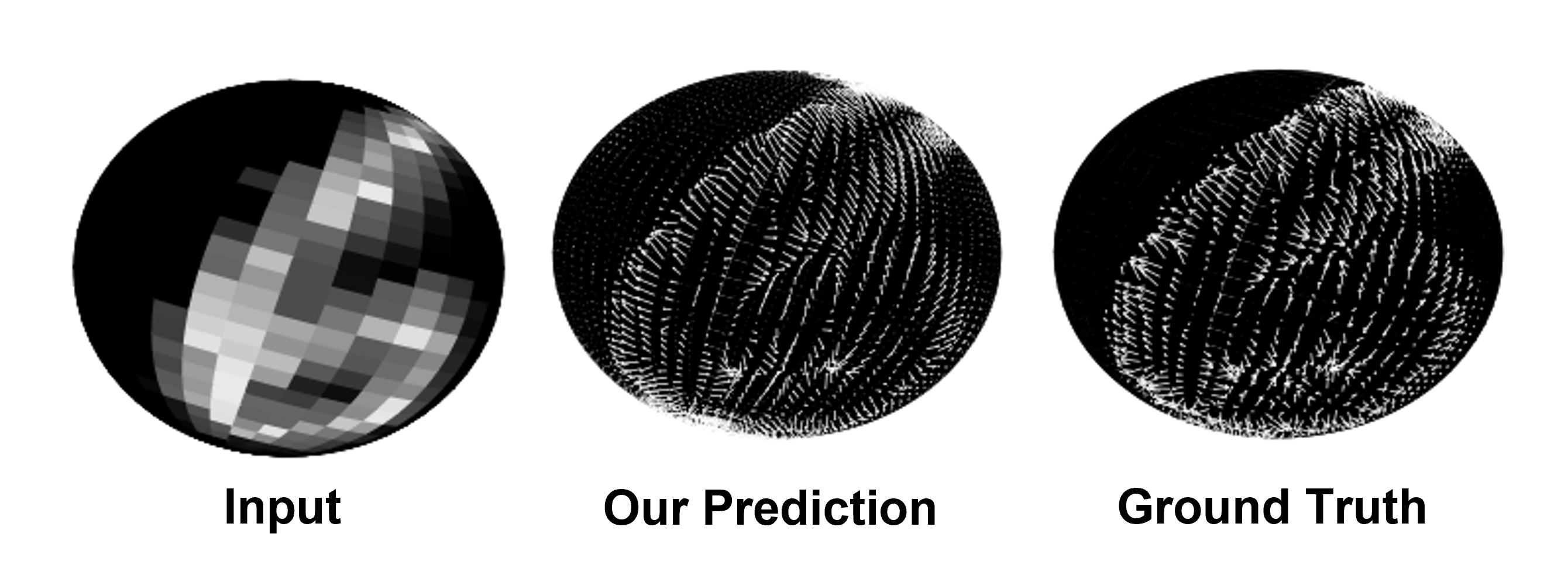}}
}
\caption{The input, output, and ground truth of the vector field prediction task.}
\label{viz_sph}
\end{figure}

We experimentally study equivariant vector field prediction on the spherical vector field MNIST (SVMNIST), a dataset proposed in~\cite{esteves2020spin}. 
We build a U-Net structure which takes the grayscale spherical image as input, and outputs a vector field on the sphere. The prediction target corresponds to the image gradients of the MNIST characters when they are mapped to a sphere as shown in Figure \ref{viz_sph}. 
We now explain the key components (convolution layer and activation layer) of our U-Net.

\begin{figure}[ht]
\centering
\scalebox{0.8}{
\centerline{\includegraphics[width=\columnwidth]{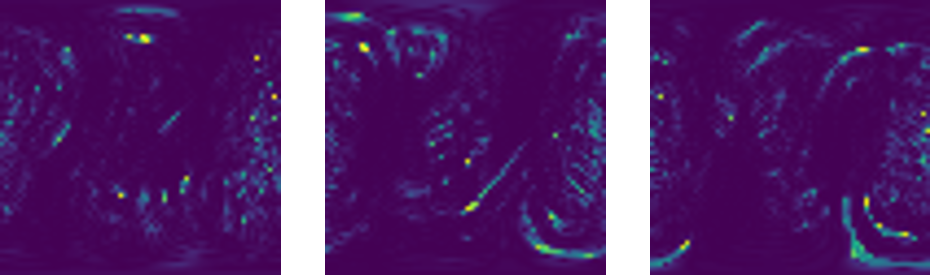}}
}
\caption{Direction change of nonzero-order features in the activation layer. The intensity in the figure reflects the angle difference before and after the activation.}
\label{sph_dir}
\end{figure}

\textbf{Convolution Layer}: 
We implement convolution in the spectral domain and parameterize the kernel via its Fourier coefficients. 
As illustrated in Section \ref{example_so3}, we implement a fast Fourier Transform that integrates over $\mathbb{S}^2$. For any hidden layers in the U-Net, we use  features of  0th and 1st order, both with the same number of channels, to increase expressivity. 
This is different from spherical convolution methods that only process features of order 0 \cite{cohen2018spherical},  and  computationally more efficient than convolution directly on $SO(3)$ \cite{cohen2018spherical}. 
\citet{esteves2020spin} define and implement convolution in the spectral domain through spin-weighted spherical harmonics, which is equivalent to the linear part in our method. 
In our paper, this method is a natural implication of the spectral sparsity, shown for all groups where a Fourier transform exists, and not just for $SO(3)$.

\textbf{Activation Layer}: We first lift the fields to the Mackey function through $f\!\!\uparrow^G(\alpha,\beta,\gamma)=e^{-im\gamma}f(\alpha,\beta)$. Then, we sum up the Mackey functions corresponding to different fields. These two steps reduce the computation compared to \cite{cohen2018spherical} due to the sparsity in the Fourier coefficients. 
The element-wise nonlinearity on features is implemented over $SO(3)$. The activated group function can then be projected back to the homogeneous space by convolution over the Fourier domain, as shown in the spectral part of figure \ref{nonlinear_part}, bringing expressivity.
Alternatively, we can pick a specific column to output for the Fourier transform, which reduces computation. In the U-Net, we always take the latter approach for better efficiency.

We report the vector prediction mean-squared errors \cite{esteves2020spin}  weighted by the spherical map sampling area. 
Our U-Net uses a number of parameters similar to the  baselines. 
The comparison is shown in Table~\ref{tab:sph}. Our network outperforms the state-of-the-art equivariant networks, especially when the input has no rotation augmentation and the testing data is rotated (NR/R). 
In addition to  being equivariant for different feature types, our model also has a more expressive nonlinearity layer. 
The novel nonlinearity entangles different fields and directional information, and can change the direction of the nonzero-order tensor in the activated output, making it more expressive than performing nonlinearity over the invariant norm \cite{esteves2020spin}. 
Figure ~\ref{sph_dir} shows the angle difference of the vector (order-one) feature maps before and after activation in a hidden layer.
Please see App.~\ref{ap:sph_details} for more details.

\begin{table}[th!]
\centering
\scalebox{0.8}{
\begin{tabular}{@{}lccc@{}}
\toprule
Method         & NR/NR        & R/R          & NR/R         \\ \midrule
Planar~\protect{\small\cite{esteves2020spin}}         & \textbf{0.3}          & 5.0          & 9.3          \\
SphCNN~\protect{\small\cite{esteves2018learning}}        & 9.7          & 31.0         & 45.6         \\
SWSCNN~\protect{\small\cite{esteves2020spin}}        & 2.9 & 3.4          & 4.3          \\
Ours           & 2.9 & \textbf{3.2} & \textbf{3.8} \\ \bottomrule
\end{tabular}
}
\caption{Results of spherical scalar to vector prediction. We report the mean-squared error $\times 10^3$ (lower is better). The baseline methods are planar (convolutional 2D CNNs), SphCNN~\cite{esteves2018learning} and SWSCNN~\cite{esteves2020spin}. NR/NR is nonrotated train and test set; NR/R is nonrotated train set and rotated test set; R/R is rotated train and test set.}
\label{tab:sph}
\end{table}

\subsection{SE(3) Prediction}
\label{sec:exp_se3}
Designing SE(3) equivariant networks for point sets is an important problem in many application areas, like chemistry and computer vision 
The standard method in the current literature is Tensor Field Networks (TFN)~\cite{thomas2018tensor}. Following our theory, we enhance 
TFN with a novel non-linearity that can capture the directional information of higher order features. 
We verify its effectiveness on two tasks: (i) QM9~\cite{ramakrishnan2014quantum} missing atom prediction 
and (ii) ModelNet40~\cite{wu20153d} point cloud shape classification.  

\textbf{Convolution Layer}: For $SE(3)$, the homogeneous space is $\mathbb{R}^3$. Since $SE(3)$ has a semidirect product structure, 
and the Fourier Transform is computationally expansive, 
we analytically derive the kernel for the convolution in the spatial domain, 
and implement convolution on the homogeneous space $\mathbb{R}^3$. As proved in the sparsity of the $SE(3)$-spectrum, the derived kernel is equivalent to that from \cite{thomas2018tensor} and \cite{weiler20183d}; 
thus, we use the convolution in TFN as implemented in~\cite{e3nn}). 

\textbf{Activation Layer}: The main difference between our extended model and the original TFN~\cite{thomas2018tensor} is the nonlinearity. After lifting the field from $\mathbb R^3$ to $SE(3)$, we project it to $\mathbb{S}^2$ attached to every point (using bandwidth $8$) to save memory,
while keeping equivariance (see App.~\ref{appendix:projection_detail} for details). To obtain more expressivity, we apply a small per-point MLP to the group function following \cite{poulenard2021functional}, which also maintains equivariance. Finally, our implementation of the projection from $SE(3)$ to $\mathbb R^3$ is equivalent to that in \cite{poulenard2021functional}, as shown in Sec.~\ref{acti}. 

\subsubsection{SE(3): Molecular Structure Completion}
\label{sec:exp_mo}
To show advantages beyond scalar field prediction, we conduct experiments to predict missing atoms in molecules. 
Here, positional prediction requires equivariant vector prediction.
The setup of the experiment follows \cite{thomas2018tensor}. 
During training, we randomly remove an atom from the molecule, and use our model to predict the atom type (order-0) and position (order-1) of the atom. 
During testing, we iteratively remove all atoms one at a time for every molecule. 
We use two metrics for evaluation. The distance error is the average error of the predicted position from the ground truth atom position; the accuracy is the proportion of the instances correctly predicted and with a distance error less than 0.5\AA. 
More details are in App.~\ref{ap:mo_details}.

We report the accuracy and distance MAE in Table~\ref{tab:molecule}. Since rotations exist naturally in molecule structures, the result illustrates the equivariance of our model. 
Our model generalizes well to test datasets of molecules with different numbers of atoms, and outperforms tensor field networks on every test dataset, which shows the effectiveness of our activation compared to the norm activation in the TFN.

 \begin{table}[ht]
   \centering
\scalebox{0.8}{
   \begin{tabular}{ccccc}
     \toprule
 &\multicolumn{2}{c} {Accuracy↑(\%)} & \multicolumn{2}{c}{Distance↓($\AA$)} \\
Atoms & TFN&Ours&TFN&Ours\\
\midrule
 19 & 93.9&\textbf{98.0}&0.14&\textbf{0.06}\\
  23&96.5&\textbf{97.1}&0.13&\textbf{0.10}\\
25-29&97.3&\textbf{98.3}&0.16&\textbf{0.10}\\
     \bottomrule
   \end{tabular}
}
   \caption{Results for missing atom prediction. We have three test datasets with 5-18, 23 and 25-29 atoms in one molecule, respectively. Every dataset has 1000 molecules.}
   \label{tab:molecule}
 \end{table}

\subsubsection{SE(3): ModelNet40 classification}
\label{sec:exp_cls}

Using our novel activation, we build a point cloud classification network based on the approaches in \cite{poulenard2021functional,thomas2018tensor}.
We report the classification accuracy in Table~\ref{tab:cls}. 
The dataset we use for the 3D object shape recognition is Modelnet40 \cite{wu20153d}, 
and it consists of 12311 3D shapes  (9843 for training and 2468 for test) over 40 categories.
Please see App.~\ref{ap:exp_cls} for more details. Our method has  performance similarly to state-of-the-art methods.

\begin{table}[ht]
\centering
\scalebox{0.8}{
  \begin{tabular}{cccc}
    \toprule
    Methods & $z/z$ & $z/\mathrm{SO}(3)$ & $\mathrm{SO}(3)/\mathrm{SO}(3)$ \\
    \midrule
    Spherical-CNN  & 88.9 & 76.7 & 86.9 \\
    $a^3$S-CNN & 89.6 & 87.9 & 88.7 \\
    SFCNN & \textbf{91.4 }& 84.8 & 90.1 \\
    TFN & 88.5 & 85.3 & 87.6 \\
    RI-Conv & 86.5 & 86.4 & 86.4 \\
    SPHNet & 87.7 & 86.6 & 87.6 \\
    ClusterNet & 87.1 & 87.1 & 87.1 \\
    GC-Conv & 89.0 & 89.1 & 89.2 \\
    RI-Framework & 89.4 & 89.4 & 89.3 \\
    VN-PointNet & 77.5 & 77.5 & 77.2 \\
    VN-DGCNN & 89.5 & 89.5 & \textbf{90.2} \\
    TFN[mlp]-P & 89.7 & \textbf{89.7} & 89.7 \\
    \midrule
    Ours & 89.7 & \textbf{89.7} & 89.3 \\
    \bottomrule
  \end{tabular}
}
  \caption{Classification accuracy in three train/test setups. Here $z$ stands for aligned data augmented by random rotations around the vertical axis and $\mathrm{SO}(3)$ indicates augmentation by random rotations. The quantitative results of previous methods are from~\cite{deng2021vector}. Please see Table~\ref{tab:cls_full} for the source of each method.
  }
  \label{tab:cls}
\end{table}

\section{Limitation and Discussion}

Several interesting directions remain for future exploration.
Although our theory covers many groups of interest for a wide range of applications, 
groups with trivial stabilizers, among others, 
and groups without Fourier transforms, 
remain unexplored. 
The input and output homogeneous spaces of each layer in this paper are the same, 
and the setting with different homogeneous spaces (different stabilizer groups) needs further study.
Another future direction is to use our unified architecture for other groups like finite permutation groups. 

\section{Conclusions}
This paper provides a Fourier perspective for  group convolutional neural networks on homogeneous spaces. We discovered a form of spectral sparsity and used it in  designing the kernel and the nonlinearity in a unified way.
Our networks showed their effectiveness in several tasks. 

\section*{Acknowledgements}
We gratefully acknowledge financial support by the grants NSF TRIPODS 1934960, ARO MURI W911NF-20-1-0080, ONR N00014-17-1-2093, and ARL DCIST CRA W911NF-17-2-0181. We also thank Carlos Esteves and Adrien Poulenard for their help, and reviewers for constructive feedback.

\newpage
\bibliography{example_paper}
\bibliographystyle{icml2022}

\newpage
\appendix
\onecolumn
 \section{Preliminary}
\subsection{Group actions and homogeneous spaces}
\label{appendix:twist}

Consider a group $G$ acting on the homogeneous space $X$,
and let $x_0$  be the origin of $X$ inducing the set of group elements $H=\{h \in G| hx_0=x_0\}$ that leaves $x_0$ unchanged, i.e., the stabilizer subgroup $H$ of $x_0$ in $G$. The set of left cosets $gH:=\left\{gh| h\in H\right\}$ is called the left quotient space $G/H$, and is isomorphic to the homogeneous space $X$. 
Take $G=SE(2)$ as the acting group and $X=\mathbb{R}^2$ as its homogeneous space. 
Any element in $g\in SE(2)$ can be denoted as $g=(t_{x},r_{\theta})$, 
where $t_x$ is the translation by the vector $x \in \mathbb{R}^2$, and $r_{\theta}$ is the rotation by the angle $\theta \in \left[0, 2\pi\right)$. 
Any rotation $r_{\theta}$ leaves the point $x_0=\left(0,0\right) \in X$ unchanged, and these rotations compose the stabilizer subgroup $H=SO(2)$. 
For any element $g=(t_x,r_{\theta}) \in SE(2)$, $gH=\left\{(t_x,r_{\theta}r_{\theta'}))| r_{\theta'} \in SO(2) \right\}=\left\{(t_x,\star) \right\}$ and all elements in the left coset $gH$ map $x_0$ to $x$. The left quotient space $G/H=\left\{x\in \mathbb{R}^2:(t_x,\star)\right\}$ is isomorphic to the homogeneous space $\mathbb{R}^2$.

The group $G$ can be viewed as a principal bundle through the partition of the group into cosets. The base space is $G/H$, and the canonical fiber is $H$, with the projection map $p: G \rightarrow G/H$, $p(g)=gH=x$ and the section $s: G/H \rightarrow G$ such that $p\circ s =id_{G/H}$, the identity map on $G/H$. 
The action of $G$ induces a twist of the fibers as $gs(x)=s(gx)\text{h}(g,x)$ where $\text{h}: G \times G/H \rightarrow H$ is the twist function. For simplicity, we denote $\text{h}(g, eH)$ as $\text{h}(g)$. 
When the group $G$ is a semidirect product group $G/H \rtimes H$, $\text{h}$ does not depend on the choice of $x$ in the homogeneous space and can be simplified to $\text{h}(g,x)=\text{h}(g)$. We use $SE(2)$ as an example to illustrate the bundle structure where the base space is $\mathbb{R}^2$ and the fiber is $SO(2)$.
Then the projection map $p: SE(2) \rightarrow \mathbb{R}^2$ is $p((t_x,r_{\theta}))=x$ and the section $s: \mathbb{R}^2 \rightarrow SE(2)$ is $s(x)=(t_x,r_0)$. For any $g=(t_x,r_{\theta}) \in SE(2)$ and any $x' \in \mathbb{R}^2$, we have $(t_x,r_{\theta})(t_{x'},r_0)=(t_{x+r_{\theta}x'},r_0)(0,r_{\theta})$, therefore the twist function $\text{h}: SE(2) \times \mathbb{R}^2 \to SO(2)$ is $\text{h}((t_x,r_{\theta}),x')=\text{h}((t_x,r_{\theta}))=r_{\theta}$.

\subsection{Irreducible Representations}
\label{appendix:irrep}
Let $V$ be a vector space over a field $K$. 
A representation of a group  G on $V$ is a homomorphism $\rho: G \rightarrow GL(V)$, i.e., for any $g_1,g_2 \in G$, $\rho(g_1g_2)=\rho(g_1)\rho(g_2)$, where $GL(V)$ is the general linear group over $V$. 

If the subspace $W$ of $V$ is invariant under the action of all group elements, that is, for any $g \in G$ and any $w \in W$, we have $\rho(g)w \in W$, we call it a sub-representation of $\rho$. 
If $\rho$ has only  two sub-representations,
the whole space $V$ and $\{0\}\subset V$, then $\rho$ is called an irreducible representation. 
If for any $g \in G$, $\rho(g^{-1})= \overline{\rho(g)^\top }$, the representation $\rho$ is called a unitary representation. Every locally compact group has a unitary representation. 

Let us denote by $U$ a unitary irreducible representation. 
In any particular basis, we can view this as a matrix, and denote by $U_{ij}$ its entry in row $i$ and column $j$.
Irreducible representations satisfy the group orthogonality theorem;
any two unitary irreducible representations $U^{l_1}$ and $U^{l_2}$ satisfy  $\left\langle U^{l_1}_{m_1n_1}, U^{l_2}_{m_2n_2}\right\rangle=\delta(l_1-l_2)\delta(m_1-m_2)\delta(n_1-n_2)$, where $\left\langle U^{l_1}_{m_1n_1}, U^{l_2}_{m_2n_2}\right\rangle$ is the inner product: $\int_{G} \overline{U^{l_1}_{m_1n_1}(g)}U^{l_2}_{m_2n_2}(g)dg$. 

Two representations $\rho_1$ and $\rho_2$ are equivalent when there exists an invertible matrix $Q$ such that for any $g\in G$,   $Q^{-1} \rho_1(g)Q=\rho_2(g)$. For a compact group or a semisimple Lie group, any representation 
$U$ 
on a Hilbert space (thus any finite-dimensional representation)
is equivalent to the direct sum of unitary irreducible representations,
that is, there exists an invertible matrix $Q$ such that $U=Q^{-1}\bigoplus_i U^iQ$, where $(U^i)_{i\in I}$ are irreducible representations of $G$ (indexed by some set $I$ that is suppressed for brevity) and $\bigoplus_i U^i$ is a block diagonal matrix with blocks $U^i$, $i\in I$. For details, we refer to (\citet{folland2016course}, Ch.~3).

For a group $G$ and a subgroup $H$, and for any representation $\rho: G \rightarrow GL(V)$ of $G$, a restricted representation $\rho|_H: H \rightarrow GL(V)$ is the restriction of $\rho$ to $H$, namely $\rho|_H(h)=\rho(h)$. Even when $\rho$ is irreducible, $\rho|_H$ may still be reducible.

Given a representation $\rho$ of the group $G$, the dual representation $\overline{\rho}$ is defined by $\overline{\rho}(g)=(\rho(g^{-1}))^{\top}$. When $\rho$ is a unitary representation, $\overline{\rho}$ is the complex conjugate of $\rho$. 
The dual representation $\overline{\rho}$ may not be equivalent to the the representation $\rho$. 
For example, for $SO(2)$ the representation $\theta\mapsto e^{im\theta}$ is not equivalent to  $\theta\mapsto e^{-im\theta}$ unless $m=0$. However, the irreducible representations are self-dual for some groups, including $SO(3)$ and the special unitary group of order two, $SU(2)$ (consisting of two-by-two complex-valued unitary matrices having unit determinant, with the multiplication operation) (\citet{fulton2013representation}, Ch.~8).

\subsection{Unimodular Separable Locally Compact Groups of Type I} 
\label{appendix:def}
 A group $G$ is a unimodular group if its left Haar measure is also a right Haar measure (\citet{folland2016course} Ch.~2). 
 A topological group is separable (second countable) 
 if its topology has a countable base (\citet{halmos2013measure}, Ch.~0). 
 If the underlying topology of the group is locally compact and Hausdorff, we call the group locally compact (\citet{stroppel2006locally}, Ch.~2). A group is said to be of Type I if each of its primary representations (factor representations) is a direct sum of copies of some irreducible representation (\citet{folland2016course}, Ch.~7). Following (\citet{gross1978evolution}, \citet{folland2016course}, Ch.~7), the Fourier Transform is well defined for any unimodular separable locally compact group of Type I.

\subsection{Clebsch-Gordan Decomposition}
\label{appendix:CG_Decomp}
 The dual group $\hat{G}$ is finite or countable for finite or compact $G$, respectively.
 For such groups, since $U_{m_1n_1}(g,p_1)U_{m_2n_2}(g,p_2) \in \mathcal{L}^2(G)$ for any $p_1, p_2 \in \hat{G}$, we can apply the Inverse Fourier Transform to express
 {\small \begin{equation*}
 U_{m_1n_1}(g,p_1)U_{m_2n_2}(g,p_2)\notag\\
    =\sum_{p \in \hat{G},\, m,n}\mathcal{C}^{p,m,n}_{p_1,p_2,m_1,m_2,n_1,n_2}U_{mn}(g,p)
\end{equation*}}
 where $m_1,m_2,n_1,n_2,m,n$ are the appropriate row and column indices. The Clebsch-Gordan coefficients $\mathcal{C}^{p,m,n}_{p_1,p_2,m_1,m_2,n_1,n_2}$ are the Fourier coefficients  $\mathcal{F}\left(U_{m_1n_1}(\cdot,p_1)U_{m_2n_2}(\cdot,p_2)\right)_{mn}(p)$.
 This so-called Clebsch-Gordan decomposition describes the decomposition of the tensor product of two irreducible representations. For $SO(3)$ and $SU(2)$, it can be shown (\citet{chirikjian2001engineering}, Ch.~10). that $\mathcal{C}^{p,m,n}_{p_1,p_2,m_1,m_2,n_1,n_2}=\mathcal{C}^{p,m}_{p_1,m_1,p_2,m_2}\mathcal{C}^{p,n}_{p_1,n_1,p_2,n_2}$,
 where $\mathcal{C}^{p,m}_{p_1,m_1,p_2,m_2}$ and $\mathcal{C}^{p,m}_{p_1,m_1,p_2,m_2}$ are the Clebsch-Gordan coefficients as they appear in \cite{kondor2018clebsch}.

\section{Details and Proofs}
\subsection{Proof of Lemma \ref{decomp}}
\label{appendix:lemma}
Since $H$ is the subgroup of $G$, $U(\cdot,p)|_H$ is a representation of $H$, but not necessarily an irreducible one. Because $H$ is a compact Lie group, there is an invertible $Q$ such that $Q^{-1}U(h,p)Q= \oplus_{i \in \mathcal{Q}(p)} \rho^i(h)$, where $\{\rho^i(h)\}_{i\in \mathcal{Q}(p)}$are the irreducible representations of $H$. 
Since $Q^{-1}U(\cdot,p)Q$ is a unitary irreducible representation of $G$, for simplicity, we will use $U(\cdot, p)$ to denote $Q^{-1}U(\cdot,p)Q$. A unitary irreducible representation of $G$ can be expressed as $U(g,p)=U(s(gH)\text{h}(g),p)$ where $s$ is the section map and $\text{h}$ is the twist function defined in App.~\ref{appendix:twist}. 
Since $U(\cdot,p)$ is a representation, based on the previous properties, $U(g,p)=U(s(gH),p)\cdot U(\text{h}(g),p)$. Thus, the unitary irreducible representation $U(\cdot,p)$ can be decomposed as:
\begin{align}\label{u-d}
    U(g,p)=U(s(gH),p)\cdot \oplus_{i \in \mathcal{Q}(p)}\rho^i(\text{h}(g)).
\end{align}

 \subsection{Proof of Proposition \ref{Fourier sparse}}
 \label{appendix:prop}

We know that $\rho$ is an unitary irreducible representation of the stabilizer subgroup $H$. 
Denote its type as $i$, writing $\rho=\rho^i$.
Writing the definition of a Mackey function in a block form, we have  $(f\!\!\uparrow^G)_k(g) = \sum_{t} \rho^i_{kt}(\text{h}(g)^{-1}) f_t(gH)$.
Using also \eqref{u-d}, 
and decomposing the integral over $G$ into integrals over $G/H$ and $H$,
the Fourier Transform of the $k$-th element 
of the vector function $f\!\!\uparrow^G$ can be calculated as follows: 
\begin{align*}
&(\widehat{(f\!\!\uparrow^G)_k})_{mn}(p)
=\int_G (f\!\!\uparrow^G)_k(g)\overline{U_{mn}(g,p)}dg
=\sum_{t,j}\int \rho^i_{kt}(\text{h}(g)^{-1}) f_t(gH)\overline{U_{mj}(s(gH),p)U_{jn}(\text{h}(g),p)}dg\\
&=\sum_{t,j}\int_{G/H} f_t(x)\overline{U_{mj}(s(x),p)}dx
\int_H \rho^i_{kt}(h^{-1})\overline{U_{jn}(h,p)}dh.
\end{align*}
We refer Theorem 2.49 in \cite{folland2016course} for the measures used in the above equation.
Since $U(h,p)=\oplus_{o \in \mathcal{Q}(p)}\rho^o(\text{h}(g))$, we known that $U_{jn}$ is either zero or belongs to a nonzero block of $U(h,p)$. When $U_{jn}$ is zero, then $\rho^i_{kt}(h^{-1})\overline{U_{jn}(h,p)}dh$ is zero. 
Else, suppose that $U_{jn}$ is an element of the irreducible representation $\rho^{f_1^p(j,n)}$, where $f_1^p$ is a function parametrized by $p$---i.e., the block $U_{jn}$ belongs to can depend on $j,n$ and $p$. For the same reason, we have $U_{jn}(h,p)=\rho^{f_1^p(j,n)}_{f_2^p(j)f_3^p(n)}(h)$.

The second integral thus becomes 
$\int_H \overline{\rho^i_{tk}(h)\rho^{f_1^p(j,n)}_{f_2^p(j)f_3^p(n)}}(h)dh$.

As mentioned in App.~\ref{appendix:CG_Decomp}, the compact group $H$ has the Clebsch-Gordan decomposition
\begin{align*}
    \rho^{l_1}_{m_1n_1}\rho^{l_2}_{m_2n_2}=\sum_{l,m,n}\mathcal{C}^{l,m,n}_{l_1,m_1,n_1,l_2,m_2, n_2}\rho^l_{mn}.
\end{align*}
Since $\left\langle \rho^l_{mn}, \rho^0_{00} \right\rangle = \delta(l)\delta(m)\delta(n)$,
the integral  $\int_H \overline{\rho^l_{tk}(h)\rho^{f_1^p(j,n)}_{f_2^p(j)f_3^p(n)}(h)}dh$ 
can only be nonzero
when the decomposition of $\rho^l_{tk}(h)\rho^{f_1^p(j,n)}_{f_2^p(j)f_3^p(n)}$ includes the trivial representation $\rho^0_{00}$.

Since $H$ is a compact Lie group, by calculating the character of the tensor product of the two irreducible representations $\rho^i$ and $\rho^{f_1^p(j,n)}$ (see App. \ref{appendix:character}), $\mathcal{C}^{00}_{l_1,m_1,n_1,l_2,m_2,n_2}$ is nonzero only when $\rho^{f_1^p(j,n)}$ is equivalent to the dual representation of $\rho^{i}$. This finishes the proof.

\subsection{Character of the Tensor Product of Two Irreducible Representations }
 \label{appendix:character}
 Suppose $\rho'$ and $\rho''$ are two unitary irreducible representations (irreps) of a compact group $G$. 
 The character of the tensor product of these two irreps is $\mathcal{X}_{\rho'\bigotimes\rho''}=\textnormal{tr}(\rho'\bigotimes\rho'')=\textnormal{tr}(\rho')\textnormal{tr}(\rho)=\mathcal{X}_{\rho'}\mathcal{X}_{\rho''}$.

 From the Inverse Fourier Transform and the properties of compact groups, we have $\rho'\bigotimes\rho''=Q^{-1}(\bigoplus \rho) Q$, where $Q$ is an invertible matrix. 
 Then $\mathcal{X} _{\rho'}\mathcal{X}_{\rho''}=\sum_{\rho}c^{\rho}_{\rho', \rho''}\mathcal{X}_{\rho}$, where $c^{\rho}_{\rho',\rho''}$ are positive integers.

The trivial representation is contained in $\rho \bigotimes \rho''$ only when $\rho$ and $\rho''$ are dual representations. 
This is because $c^{\rho^0}_{\rho',\rho''}=\int\mathcal{X}_{\rho'}\mathcal{X}_{\rho''}=\int\mathcal{X}_{\rho'}\overline{\mathcal{X}_{\overline{\rho''}}}=\left\langle\mathcal{X}_{\rho'}, \mathcal{X}_{\overline{\rho''}}\right\rangle$. The integral can be nonzero only when $\overline{\rho''}$ and $\rho$ are equivalent, where $\overline{\rho''}$ is the dual representation of the unitary irreducible representaion $\rho''$.

\subsection{Proof of Corollary \ref{ker-sparse}}
 \label{appendix:coro}
Due to the convolution theorem (\citet{chirikjian2001engineering}, Ch.8), 
we have $\widehat{f_2\!\!\uparrow^G}(p)=\widehat{f_1\!\!\uparrow^G}(p)\hat{\kappa}(p)$. The sparsity pattern exists in $\widehat{f_1\!\!\uparrow^G}(p)$ and $\widehat{f_2\!\!\uparrow^G}(p)$ due to Proposition \ref{Fourier sparse}.
Suppose $\widehat{f_1\!\!\uparrow^G}(p)$ has zeroes in columns $A_1$, and ${f_2\!\!\uparrow^G}(p)$ has zeroes columns columns $A_2$. Then
the equation  $\widehat{f_2\!\!\uparrow^G}(p)=\widehat{f_1\!\!\uparrow^G}(p)\hat{\kappa}(p)$
can be reduced by taking the subsets of rows in the complement of $A_1$ and columns  in the complement of $A_2$.
It is easy to see that the entries of $\hat{\kappa}$ outside of these indices do not enter the calculation. Thus, they can be taken to be anything, and in particular as zeroes.
This proves the desired claim.

\subsection{Discussion of completeness}
 \label{appendix:discussion of completeness}
Revisiting $\int_H \rho^{l_1}_{m_1n_1}(h)\rho^{l_2}_{m_2n_2}(h) dh$, if the integral is nonzero, then $\rho^{l_2}$ is equivalent to $\overline{\rho^{l_1}}$. Supposing $\rho^{l_2}=Q\overline{\rho^{l_1}}Q^{-1}$, we have 
\begin{align*}
    &\int_H \rho^{l_1}_{m_1n_1}(h)\rho^{l_2}_{m_2n_2}(h) dh=\int\sum_{a,b} \rho^{l_1}_{m_1n_1}Q_{m_2a}\overline{\rho^{l_1}}_{ab}Q^{-1}_{bn_2}dh\\
    &=\sum_{a,b}Q_{m_2a}Q_{n_2b}\int_H\rho^{l_1}_{m_1n_1}\overline{\rho^{l_1}}_{ab}dh
    =Q_{m_2m_1}Q_{n_2n_1}.
\end{align*}

Therefore, the value of the integral is related to $Q$. 
We  can choose a unitary representation of $G$, and a basis of the underlying vector space, such that $\rho^{l_2}$ is the dual representation of $\rho^{l_1}$.
This holds because the representation of compact Lie groups can split into an orthogonal direct sum of irreducible finite-dimensional unitary representations (according to the Peter-Weyl Theorem) and the dual representation of a finite-dimensional representation is irreducible.

In this case, when $\widehat{(f\!\!\uparrow^G)}$ is a Mackey function  lifted via $f\!\!\uparrow^G(g)=\rho(\text{h}(g)^{-1})f(gH)$ from a field $f$, $\widehat{(f\!\!\uparrow^G)_k}$ is nonzero only on the $k$-th column in the corresponding block and 
$\widehat{(f\!\!\uparrow^G)_k}_{mn}=\widehat{(f\!\!\uparrow^G)_{k+1}}_{m(n+1)}$ for any $k,m,n$. The converse also stands.

For the convolution $(f_{out}\!\!\uparrow^G)_a(g)=\int_G\sum_b\kappa(\nu^{-1}g)_{ab}(f_{in}\!\!\uparrow^G)_b(\nu)$, in the Fourier domain we have $\widehat{(f_{out}\!\!\uparrow^G)_a}_{mn}=\sum_{b,t}\widehat{(f_{in}\!\!\uparrow^G)_b}_{mt}\widehat{\kappa_{ab}}_{tn}$. 
Now, $f_{out}$ obeys the above property and $f_{in}$ is arbitrary, therefore, we have $\widehat{\kappa_{ab}}_{mn}=\widehat{\kappa_{(a+1)b}}_{m(n+1)}$ for any $a,b,m,n$. 
Meanwhile, the nonzero columns in  $\widehat{\kappa_{ab}}$ and $\widehat{(f_{out}\!\!\uparrow^G)_a}$ 
have the same indices, 
which implies that $\kappa(gh)=\rho_{out}(h^{-1})\kappa(g)$.  

On the other hand, $\widehat{(f_{in}\!\!\uparrow^G)}$ is also a Mackey function, and the relation between its Fourier coefficients can help us simplify $\widehat{\kappa}$. For any $\kappa$ with the above constraint, we can design a kernel $\kappa^*$ that satisfies the following constraints:

 1. $\widehat{\kappa^*_{ab}}_{mn}=\widehat{\kappa^*_{a(b+1)}}_{(m+1)n}$ for any $a,b,m,n$;
 
 2. the nonzero rows in $\widehat{\kappa^*_{ab}}$ and the nonzero columns in $\widehat{(f_{in}\!\!\uparrow^G)_b}$ have the same indices;
 
 3. the nonzero columns in 
 $\widehat{\kappa^*_{ab}}$ and $\widehat{(f_{out}\!\!\uparrow^G)_a}$ have the same indices, and $\widehat{\kappa^*_{ab}}_{mn}=\widehat{\kappa^*_{(a+1)b}}_{m(n+1)}$ for any $a,b,m,n$.
 
such that $\int_G\kappa(\nu^{-1}g)(f_{in}\!\!\uparrow^G)(\nu)=\int_G\kappa^*(\nu^{-1}g)(f_{in}\!\!\uparrow^G)(\nu)$. 
Meanwhile, from the Fourier domain, we know that when $\kappa_1^* \neq \kappa_2^*$, then $\int_G\kappa_1^*(\nu^{-1}g)(f_{in}\!\!\uparrow^G)(\nu)\neq\int_G\kappa_2^*(\nu^{-1}g)(f_{in}\!\!\uparrow^G)(\nu)$. Therefore we find a complete characterization of the equivariant linear map $\left\{\kappa^*\right\}$.

In spatial domain, since $ \int \kappa_{st}(g)\overline{U_{mn}(g)} dg=\int \kappa^\top _{ts}(g^{-1})\overline{\overline{U_{nm}(g)}}dg$, we know that $f(g)=\kappa^\top (g^{-1})$ is also a Mackey function, 
satisfying 
$f(gh)=\overline{\rho_{in}}(h^{-1})f(g)$. 
Therefore, we have $\kappa^*(hg)=\kappa^*(g)\rho_{in}(h^{-1})$. 
This is equivalent to the space of kernels such that $\kappa^*(h_1gh_2)=\rho(h_2^{-1})\kappa^*\rho(h_1^{-1})$, from \cite{weiler2019general}.  \cite{weiler2019general} proves that 
these are all equivariant kernels, which finishes our proof

\section{Details and Proofs for the Examples}
\subsection{$SE(3)$}
Consider the group $SE(3)$, with the stabilizer subgroup $SO(3)$ and the homogeneous space $\mathbb{R}^3$.
For any element $(x,R) \in SE(3)$, where $x \in \mathbb{R}^3$ and $R \in SO(3)$, the unitary irreducible representations of $SE(3)$ have the form, for $p\in \mathbb{R}^{+}$ and an integer $s$,
\begin{align}
    U^s_{l',m';l,m}(x,R;p)=\sum^l_{j=-l}[l',m'|p,s|l,j](x)\tilde{U}^l_{jm}(R),
\end{align}
where 
\begin{small}
\begin{align*}
   &[l',m'|p,s|l,j](x)=(4\pi)^{\frac{1}{2}}\sum^{l'+l}_{k=|l'-l|} i^k \sqrt{\frac{(2l'+1)(2k+1)}{(2l+1)}}J_k(pa),\\
   &\mathcal{C}(k,0;l',s|l,s)\mathcal{C}(k.m-m';l',m'|l,m) Y^{m-m'}_k(\theta, \phi),
\end{align*}
\end{small}
and $\tilde{U}^l_{mn}=(-1)^{n-m}D^l_{mn}$, where $D^l_{mn}$ is the $l$-th order Wigner $D$-matrix---an unitary irreducible representation of $SO(3)$, $\mathcal{C}$ is a Clebsch-Gordan coefficient, $Y_k$ is a $k$-th order spherical harmonic,
$J_k$ is the $k$-th order spherical Bessel function, $a$ is the length of $x$, and $\theta$, $\phi$ are the spherical coordinates of the unit vector $\hat{x}$.

The lifting process takes the form 
\begin{align*}
    f\!\!\uparrow^G(g) =f\!\!\uparrow^G(x,R)= D^l(R^\top ) f(x). 
\end{align*}

The Fourier Transform $(\widehat{f\!\!\uparrow^G(x,R)_k})^s_{l,m;l',m'}(p)$ (see App.~\ref{appendix: $SE(3)$_detail} for details) is 
\begin{small}
\begin{equation*}
     \sum_j\int_{\mathbb{R}^3} (-1)^{-k-j}f_{-j}(x,I)\overline{U^s_{l,m;l',j}(x,I;p)}dx\delta(l'-t) \delta(k+m'),
\end{equation*} 
\end{small}
where $I$ is the identity matrix, the identity element in $SO(3)$.

The convolution on $SE(3)$ takes the form
\begin{align*}
(\kappa \ast l)_i (g) = \sum_j\int_{\nu\in G} \kappa_{ij}(\nu^{-1}g)l_j(\nu) d\nu.
\end{align*}
The Fourier transform $\mathcal{F}((\kappa \ast l)_i)^s_{l',m'; l,m}(p)$ becomes
\begin{align*}
    \sum_j \sum^{\infty}_{a=|s|}\sum^a_{b=-a} (\widehat{l_{j}})^s_{l',m';a,b}(p)(\widehat{\kappa_{ij}})^s_{a,b;l,m}(p).
\end{align*}
Assume that the order of the input field is $l_1$, and the order of the output field is $l_2$. Then  $\kappa_{ij}$ only has nonzero Fourier coefficients at $\widehat{\kappa_{ij}}^s_{l_1,-j;l_2,-i}(p)$. 
Meanwhile, we can prove that $\widehat{\kappa_{ij}}^s_{l_1,-j;l_2,-i}(p)$ 
has the form  $(-1)^{j-i}c$, 
where $c$ is a constant over $i,j$, for direct convolution on $\mathbb{R}^3$ (see App\ref{appendix: $SE(3)$_kij}).
Then we can project $\kappa$ to the homogeneous space $\mathbb{R}^3$, obtaining $\kappa'_{ij}(x)=\kappa_{ij}(x,I)$ as
\begin{align*}
\label{example_se3}
    & \frac{1}{2\pi^2} \sum^{\min(l_1,l_2)}_{s=-\min(l_1,l_2)} \int^{\infty}_0 (-1)^{j-i}c^s(p)U^s_{l_1,-j; l_2,-i}(x,I;p)p^2dp
=\sum^\top _{m=-t} \sum^{l_1+l_2}_{t=|l_1-l_2|} C_t(\|x\|) \mathcal{C}^{l_2,i}_{t,m,l_1,j}\overline{Y^m_t}.
\end{align*}

We see that $\kappa$ is in the span of a fixed basis,  
and the weight function $C_t$ of $\|x\|$ is arbitrary and thus learnable.
This result is consistent with the form of tensor field networks in \citet{thomas2018tensor} and 3D steerable CNNS in \citet{weiler20183d}.

\begin{figure}[ht]
\centerline{\includegraphics[width=0.5\columnwidth]{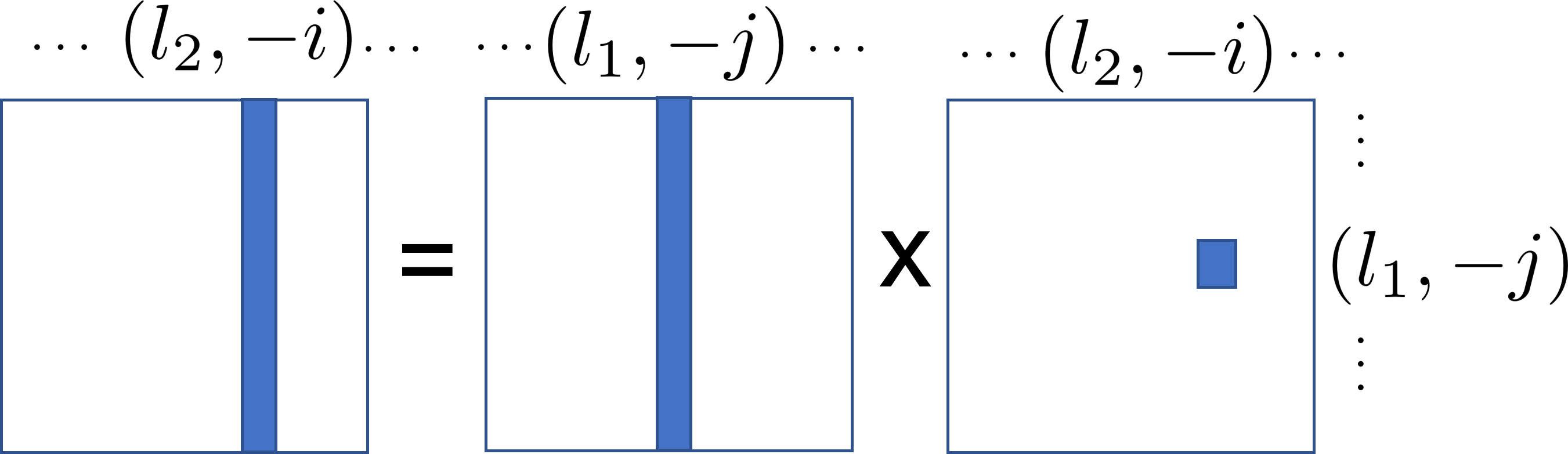}}
\caption{Fourier matrix multiplication for SE(3)}
\label{se3_mul}
\end{figure}

\subsection{Details for $SO(3)$}
\label{appendix:$SO(3)$_detail}
The Fourier Transform becomes:
 \begin{align*}
     &\widehat{f\!\!\uparrow^G}^l_{mn}=\int_{R \in SO(3)} f\!\!\uparrow^G(R) \overline{D^l_{mn}}(R) dR
     =\int_{(\alpha,\beta,\gamma) \in SO(3)} e^{-ik\gamma} f(\alpha, \beta)e^{im\alpha}d^l_{mn}(\beta)e^{in\gamma}
     d\alpha  \sin(\beta) d\beta d\gamma\\
     &=\int_{(\alpha, \beta)\in \mathbb{S}^2} f(\alpha, \beta)e^{im\alpha}d^l_{mn}(\beta)d\alpha  \sin(\beta) d\beta
    \int_{\gamma \in SO(2)} e^{-ik\gamma} e^{in\gamma} d\gamma.
 \end{align*}
This further equals
\begin{align*}
    \int_{(\alpha, \beta)\in \mathbb{S}^2} f(\alpha, \beta)e^{im\alpha}d^l_{mn}(\beta)d\alpha  \sin(\beta) d\beta\cdot 2\pi\delta(k-n).
\end{align*}
Projecting $f_{out}\!\!\uparrow^G$ to $f_{out}$ takes the form
\begin{align*}
     &f_{out}(\alpha,\beta)
     =f_{out}\!\!\uparrow^G(\alpha, \beta, 0)
     =\int_G \kappa(R_{g}^{-1}R_{\alpha, \beta ,0})f_{in}\!\!\uparrow^G(g)dg\\
     &=\int_G \kappa(R_{\alpha', \beta', \gamma'}^{-1}R_{\alpha, \beta ,0})f_{in}\!\!\uparrow^G(\alpha',\beta',\gamma')d\alpha' \sin(\beta')d\beta' d\gamma'\\
     &=\int_G \kappa(R_{z}(-\gamma')R_{y}(-\beta')R_z(-\alpha')R_z(\alpha) R_y(\beta)) e^{-im\gamma'}f_{in}(\alpha',\beta')d\alpha' \sin(\beta')d\beta' d\gamma'\\
     &=\int_G e^{im_1\gamma'}\kappa(R_{y}(-\beta')R_z(-\alpha')R_z(\alpha) R_y(\beta)) e^{-im_1\gamma'}f_{in}(\alpha',\beta')d\alpha' \sin(\beta')d\beta' d\gamma'\\
     &=C\int_{\mathbb{S}^2} \kappa(R_{y}(-\beta')R_z(-\alpha')R_z(\alpha) R_y(\beta)) f_{in}(\alpha',\beta')d\alpha' \sin(\beta')d\beta'\\
    &=C\int_{\mathbb{S}^2} e^{-im_2\text{h}(R^{-1}(
     \alpha',\beta',0)R(\alpha,\beta,0))} \kappa'(R_{y}(-\beta')R_z(-\alpha')x_{\alpha,\beta}) f_{in}(\alpha',\beta')d\alpha' \sin(\beta')d\beta'.
\end{align*}
This completes the details for $SO(3)$.

\subsection{Fourier Transform over $SE(3)$}
\label{appendix: $SE(3)$_detail}
The Fourier Transform has the form
\begin{align*}
   &(\widehat{f\!\!\uparrow^G(x,R)_k})^s_{l,m;l',m'}(p)
   = \int_{SE(3)}f\!\!\uparrow^G(x,R)_k(x,R)\overline{U^s_{l,m;l',m'}(x,R;p)}dRdx\\
   &=\sum_{g,l',j}\int_{SE(3)} D^\top _{kg}(R^\top )f_g(x,I) \overline{U^s_{l,m;l',j}(x,I,p)}
   \overline{\tilde{U}^{l'}_{jm'}(R)}dRdx\\
   &=\sum_{g,l',j}\int_{\mathbb{R}^3} f_g(x,I)\overline{U^s_{l,m;l',j}(x,I;p)}dx
   \int_{SO(3)}(-1)^{m'-j} \overline{D^\top _{gk}(R)D^{l’}_{jm'}(R)}dR.
 \end{align*}

 Due to the fact that Clebsch–Gordan coefficients
 $\mathcal{C}^{00}_{l_1,m_1,l_2,m_2}=\delta(l_1-l_2)
 \delta(m_1+m_2)$,
 the decomposition of  $D^\top _{gk}(R)D^{l’}_{jm'}(R)$ includes the trivial representation only when $t=l'$, $g=-j$ and $k=-m'$. Therefore, the Fourier matrix has sparsity pattern
 \begin{align*}
     &(\widehat{f\!\!\uparrow^G(x,R)_k})^s_{l,m;l',m'}(p)
     =\sum_j\int_{\mathbb{R}^3} (-1)^{-k-j}f_{-j}(x,I)\overline{U^s_{l,m;l',j}(x,I;p)}dx
     \quad\delta(l'-t) \delta(k+m'). 
 \end{align*}
 
 \subsection{Kernel Property for $SE(3)$}
 \label{appendix: $SE(3)$_kij}
 From the Fourier transform, we know that $\widehat{f_{out}\!\!\uparrow^G_i}^s_{l',m'; l_1,-i}(p)$ 
 has the form $(-1)^{-i}C$, where $C$ does not depend on  $i$, 
 and $\widehat{f_{in}\!\!\uparrow^G_j}^s_{l',m'; l_1,-j}(p)$ 
 has the form $(-1)^{-j}C'$, where $C'$ does not depend on $j$. 
 Thus, $\sum_{j}\widehat{\kappa_{ij}}^s_{l_1,-j;l_2,-i}(p)$ has the form $(-1)^{-i}C_j$, where $C_j$ does not depend on $i$.
 Therefore, we find the kernel $\kappa_{ij}$ whose Fourier matrix is shown in Figure \ref{se3_mul}, and $\kappa_{ij}(g)$ is
 \begin{align*}
     \frac{1}{2\pi^2} \sum^{\min(l_1,l_2)}_{s=-\min(l_1,l_2)} \int^{\infty}_0 (-1)^{-i}c^s_{l_1,-j;l_2,-i}(p)U^s_{l_1,-j; l_2,-i}(x,R;p)
 \end{align*}
 where $\sum_jc^s_{l_1,-j;l_2,-i}$ does not depend on $i$.

 To  directly apply the convolution on a homogeneous space acted on by a semidirect product, $c^s_{l_1,-j;l_2,-i}(p)$ should have the form $(-1)^{j-i}c$, where $c$ is does not depend on $i$ or $j$. 
 This is because $\kappa$ and the function $p: G \rightarrow V$, where $p(g^{-1})=\kappa^\top (g)$ for any $g \in G$, should both be Mackey functions.

\section{Experiment Details}
\subsection{Details of Spherical U-Net in Sec.~\ref{sec:exp_sph}}
\label{ap:sph_details}
The input field to the U-Net has one order-0 feature and the output of the U-Net is one order-1 feature. 
The overall network has six layers of widths $[32, 16, 16, 16, 16, 32]$. 
The hidden features have $[8, 12, 16, 12, 8]$ channels, respectively, for each type of field. 
The types in hidden features are order-0 and order-1. 
The loss is the same as ~\cite{esteves2020spin}, the mean-squared error weighted by the spherical map sampling area.
The whole network is trained end-to-end from scratch using the Adam optimizer with an initial learning rate of $1\times 10^{-3}$. The learning rate decays by a factor of $0.2$ at epochs $10$ and $15$.

\subsection{Nonlinearity in the  $SE(3)$ experiment (Sec.~\ref{sec:exp_se3})}
\label{appendix:projection_detail}
When we only take the $-1$-st, $0$-th and $1$-st element of the vector-valued function $l$ on $SE(3)$, 
the $0$-th element $l_0(*,x)$, of $l$ is a Mackey function on $SO(3)$ at every point $x$.
This is 
shown in the section \ref{acti}. 
We observe that $x\mapsto\begin{pmatrix} l_{-1}(*,x)\\ l_1(*,x)\end{pmatrix}$ is a Mackey function on $SO(3)$ corresponding to the irreducible representation $\theta\mapsto\begin{pmatrix}\cos(\theta)&  - \sin(\theta)\\ \sin(\theta)&\cos(\theta)\end{pmatrix}$ 
 of $SO(2)$,
using the real form of the Wigner-D matrix. 
Therefore we can project $x\mapsto\begin{pmatrix} l_{-1}(R,x)\\ l_1(R,x)\end{pmatrix}$ to the sphere by taking $\begin{pmatrix} l_{-1}(r,x)\\ l_1(r,x)\end{pmatrix}$ and using its norm. For $l_0(R,x)$, we only need to take $l_0(r,x)$. Through the above method, we can project the signal from $SO(3)$ to the sphere and keep the equivariance. 

\subsection{Details of Molecular Completion in Sec.~\ref{sec:exp_mo}}
\label{ap:mo_details}
Our network contains five convolution activation layers using $0,1,2$-order fields as hidden features with corresponding hidden channel dimensions $[32,32,32,32]$. In the last layer, we output six scalars for every existing atom (one for probability and five for one-hot atom type prediction for the missing atom)
and one vector (relative position to the missing atom).
The losses are the same as described in ~\cite{thomas2018tensor} and the network is trained using the Adam optimizer with the initial learning rate $1\times 10^{-3}$.
The learning rate is decreased by a factor $0.3$ at epoch $2500$.

\subsection{Details of Shape Classification in Sec.~\ref{sec:exp_cls}}
\label{ap:exp_cls}
Our network follows~\cite{poulenard2021functional, poulenard2019effective} to perform PCA aligned KD-Tree pooling. 
It contains six convolution activation layers before global pooling, with KD-Tree pooling depth factors of $[0,2,0,2,0,2]$, leading to $[1024, 1024, 256, 256, 64, 64, 16]$ points starting from the input point cloud. 
Global pooling is applied to the function lifted to the group as in~\cite{poulenard2021functional}. 
Since in this task the translation equivariance can be trivially addressed via subtracting the center of mass, following~\cite{poulenard2021functional,dym2020universality} we concatenate the global $xyz$ coordinates as an order-1 feature to each layer's input.
We use the standard cross entropy loss.
The network is trained via the Adam optimizer with a starting learning rate of $1\times 10^{-3}$. 
The learning rate is decayed by a factor of $0.3$ at epochs $[100,150,200,250,300]$.

\subsection{Table~\ref{tab:cls} with references}

\begin{table}[h!]
  \centering
  \begin{tabular}{lccc}
    \toprule
    Methods & $z/z$ & $z/\mathrm{SO}(3)$ & $\mathrm{SO}(3)/\mathrm{SO}(3)$ \\
    \midrule
    Spherical-CNN~\cite{esteves2018learning}  & 88.9 & 76.7 & 86.9 \\
$a^3$S-CNN~\cite{liu2018deep} & 89.6 & 87.9 & 88.7 \\
    SFCNN~\cite{rao2019spherical} & \textbf{91.4 }& 84.8 & 90.1 \\
    TFN~\cite{thomas2018tensor} & 88.5 & 85.3 & 87.6 \\
    RI-Conv~\cite{zhang2019rotation} & 86.5 & 86.4 & 86.4 \\
    SPHNet~\cite{poulenard2019effective} & 87.7 & 86.6 & 87.6 \\
    ClusterNet~\cite{chen2019clusternet} & 87.1 & 87.1 & 87.1 \\
    GC-Conv~\cite{zhang2020global} & 89.0 & 89.1 & 89.2 \\
    RI-Framework~\cite{li2020rotation} & 89.4 & 89.4 & 89.3 \\
    VN-PointNet~\cite{deng2021vector} & 77.5 & 77.5 & 77.2 \\
    VN-DGCNN~\cite{deng2021vector} & 89.5 & 89.5 & \textbf{90.2} \\
    TFN[mlp]-P~\cite{poulenard2021functional} & 89.7 & \textbf{89.7} & 89.7 \\
    \midrule
    Ours & 89.7 & \textbf{89.7} & 89.3 \\
    \bottomrule
  \end{tabular}
  \caption{Due to the page limit, we provide the references for the methods in Table~\ref{tab:cls} here.}
  \label{tab:cls_full}
\end{table}

\end{document}